\documentclass[acmtog]{acmart}

\setcopyright{acmcopyright}\acmJournal{TOG}
\acmYear{2021}\acmVolume{40}\acmNumber{4}\acmArticle{56}\acmMonth{8} \acmDOI{10.1145/3450626.3459676}

\usepackage{mathtools}
\usepackage{amsmath}

\usepackage{amssymb}
\usepackage{amsfonts}
\usepackage{amsopn}
\usepackage{graphicx} 
\usepackage{textcomp}
\usepackage{xfrac}
\usepackage{bbm}
\usepackage{overpic}
\usepackage{subfig}
\usepackage{wrapfig}
\usepackage[ruled,vlined,linesnumbered]{algorithm2e}
\SetAlFnt{\small}
\SetAlCapFnt{\small}
\SetAlCapNameFnt{\small}
\SetAlCapHSkip{0pt}

\usepackage{multirow}

\usepackage{booktabs}
\usepackage{footnote}
\usepackage{paralist}
\usepackage{enumitem}
\usepackage{autobreak}
\usepackage{hyperref}
\usepackage{cleveref}

\usepackage{color}
\definecolor{green}{rgb}{0, 0.4, 0}
\definecolor{orange}{rgb}{0.8, 0.6, 0.2}
\definecolor{red}{rgb}{1.0, 0.0, 0.0}
\definecolor{teal}{rgb}{0.0, 0.4, 0.4}
\definecolor{purple}{rgb}{0.65,0,0.65}
\definecolor{saffron}{rgb}{0.95,0.75,0.2}
\definecolor{turquoise}{rgb}{0.0,0.5,0.5}
\definecolor{brown}{rgb}{0.5, 0.16, 0.16}

\usepackage{overpic}
\usepackage{currfile} 

\newlength\savedwidth

\newcommand{\supl}[1]{{\color{black}#1}}

\newcommand{\kx}[1]{{\color{black}#1}}

\definecolor{lightgray}{rgb}{0.6, 0.6, 0.6}

\newcommand{\Fig}[1]{Figure~\ref{fig:#1}}
\newcommand{\Eq}[1]{Eq.~(\ref{eq:#1})}
\newcommand{\Eqs}[2]{Eq.~(\ref{eq:#1}--\ref{eq:#2})}

\newcommand{\Sec}[1]{Section~\ref{sec:#1}}
\newcommand{\Algo}[1]{Algorithm~\ref{algo:#1}}
\newcommand{\AlgoL}[1]{Line~\ref{algol:#1}}
\newcommand{\Tab}[1]{Table~\ref{tab:#1}}

\usepackage[normalem]{ulem}

\newcommand{\hidecomment}[1]{}

\newcommand{\bR}{\mathbf{R}}

\newcommand{\br}{\mathbf{r}}

\newcommand{\bx}{\mathbf{x}}
\newcommand{\bc}{\mathbf{c}}
\newcommand{\bu}{\mathbf{u}}
\newcommand{\bs}{\mathbf{s}}
\newcommand{\bv}{\mathbf{v}}

\newcommand{\bt}{\mathbf{t}}

\newcommand{\bepsilon}{{\bf \epsilon}}
\newcommand{\bT}{\mathbf{T}}

\newcommand{\bzero}{\mathbf{0}}
\newcommand{\bone}{\mathbf{1}}

\newcommand{\best}[1]{{$\mathbf{\color{blue} #1}$}}
\newcommand{\sbest}[1]{{$\mathbf{\color{green} #1}$}}

\usepackage{xspace}
\newcommand{\DSICL}{\texttt{ICL-NUIM}\xspace}
\newcommand{\DSTUM}{\texttt{TUM RGB-D}\xspace}
\newcommand{\DSETH}{\texttt{ETH3D}\xspace}
\newcommand{\DSETHCS}{\texttt{ETH3D-CS}\xspace}
\newcommand{\DSFCM}{\texttt{FastCaMo}\xspace}
\newcommand{\DSFCMS}{\texttt{FastCaMo-Synth}\xspace}
\newcommand{\DSFCMR}{\texttt{FastCaMo-Real}\xspace}
\newcommand{\DSREPL}{\texttt{Replica}\xspace}

\begin{document}
\title{ROSEFusion: Random Optimization for Online Dense Reconstruction under Fast Camera Motion}

\author{Jiazhao Zhang}
\affiliation{%
  \institution{National University of Defense Technology}
  \country{China}
}
\author{Chenyang Zhu}
\affiliation{%
  \institution{National University of Defense Technology}
  \country{China}
}
\author{Lintao Zheng}
\affiliation{%
  \institution{National University of Defense Technology}
  \country{China}
}
\author{Kai Xu}
\affiliation{%
  \institution{National University of Defense Technology}
  \country{China}
  \authornote{Corresponding author: Kai Xu (kevin.kai.xu@gmail.com)}
}

\begin{abstract}

Online reconstruction based on RGB-D sequences has thus far been restrained to relatively slow camera motions (<1m/s). Under very fast camera motion (e.g., 3m/s), the reconstruction can easily crumble even for the state-of-the-art methods. Fast motion brings two challenges to depth fusion: 1) the high nonlinearity of camera pose optimization due to large inter-frame rotations and 2) the lack of reliably trackable features due to motion blur. We propose to tackle the difficulties of fast-motion camera tracking \kx{in the absence of inertial measurements} using random optimization, in particular, the Particle Filter Optimization (PFO). To surmount the computation-intensive particle sampling and update in standard PFO, we propose to accelerate the randomized search via updating a particle swarm template (PST). PST is a set of particles pre-sampled uniformly within the unit sphere in the 6D space of camera pose. Through moving and rescaling the pre-sampled PST guided by swarm intelligence, our method is able to drive tens of thousands of particles to locate and cover a good local optimum extremely fast and robustly. The particles, representing candidate poses, are evaluated with a fitness function defined based on depth-model conformance. Therefore, our method, being depth-only and correspondence-free, mitigates the motion blur impediment as \kx{(ToF-based) depths are often resilient to motion blur.} Thanks to the efficient template-based particle set evolution and the effective fitness function, our method attains good quality pose tracking under fast camera motion (up to 4m/s) in a realtime framerate without including loop closure or global pose optimization. Through extensive evaluations on public datasets of RGB-D sequences, especially on a newly proposed benchmark of fast camera motion, we demonstrate the significant advantage of our method over the state of the arts.

\end{abstract}

\vspace{-18pt}

\vspace{-26pt}
\vspace{-28pt}

\begin{teaserfigure}
\begin{overpic}[width=1.0\textwidth,tics=5]{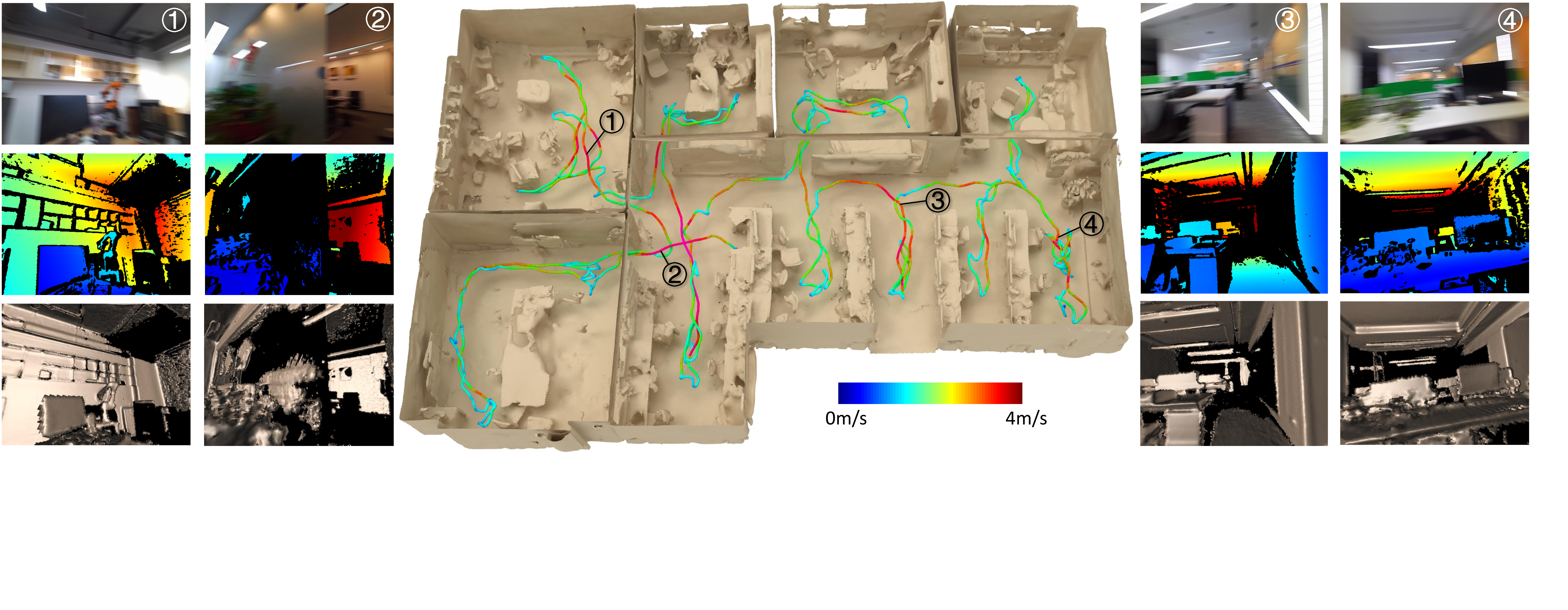}
\end{overpic}\vspace{-4pt}
\caption{
We introduce ROSEFusion, a depth-only online dense reconstruction which is stable and robust to highly fast camera motion. Built upon the volumetric depth fusion framework, our method solves the highly nonlinear optimization problem of fast-motion camera tracking using random optimization. In this example, the depth camera moves at a speed of $2$m/s in average and up to $3.6$m/s, leading to severe motion blur in RGB images. This sequence would break most state-of-the-art online reconstruction methods. Thanks to our novel particle filter optimization with swarm intelligence, our method is able to fuse the depth maps (resilient to motion blur) accurately, attaining a satisfying reconstruction quality.
See the planar walls and the correct overall layout; the imperfect local geometry was mainly due to incomplete depth scanning. The tracked camera trajectory is visualized and color-coded with camera speed.}
\label{fig:teaser}
\end{teaserfigure} 

\maketitle

\section{Introduction}
\label{sec:intro}
With the proliferation of commodity RGB-D cameras, the recent decade has witnessed a booming of online RGB-D reconstruction techniques. Since the seminal work of KinectFusion~\cite{Newcombe2011}, a huge body of works have been proposed based either on volumetric depth fusion~\cite{Niessner2013,chen2013scalable} or point-based fusion~\cite{keller2013real,Whelan2015}. Despite the significantly advanced frontier on reconstruction quality and scalability, the prior systems have hereunto been restrained to \emph{slow camera motions}, typically less than 1m/s. Under faster camera motions, the RGB-D reconstruction could easily collapse due to drastically increased difficulties in tracking and mapping. This has greatly limited the practical use of these techniques especially in autonomous scanning and reconstruction by, e.g., UAVs.

Fast camera motion brings two main challenges to RGB-D fusion. \emph{First}, fast motions lead to large rotations which make camera pose optimization highly nonlinear and prune to local optima for gradient descent methods~\cite{schmidt2001using}. \emph{Second}, fast camera movement causes severe motion blur in RGB images, \kx{especially under dark lighting conditions and in the case of indoor-level camera-to-scene distance}. This makes it intractable for \kx{robust photometric tracking} which is crucial to many SLAM~\cite{mur2015orb} and dense RGB-D reconstruction methods~\cite{Dai2017}.

We propose, ROSEFusion, the first end-to-end solution to online dense RGB-D reconstruction under \emph{fast camera motions} \kx{without inertial measurement.}
Our method relies solely on depth information for robust camera tracking and accurate volumetric fusion, observing that depth sensors (especially time-of-flight (ToF)) are usually more resilient to motion blur. When the camera moves fast, depth map defect often exhibits as overshoot or undershoot in the transition between foreground and background~\cite{hansard2012time} but not as full-frame pixel mixing/blurring as in RGB images (see \Fig{teaser}). The detection and removal of the false depth signals appearing around occlusion edges can be easily realized with hardware~\cite{lee2014time} and has usually been done by most commercial ToF cameras.

To tackle the challenges in fast-motion camera tracking mentioned above, we propose two key designs.
\emph{Firstly}, we propose to solve the highly nonlinear optimization of large inter-frame camera transformation by \emph{random optimization}, in particular, the Particle Filter Optimization (PFO)~\cite{liu2016particle}. To surmount the computationally intensive particle sampling and update in standard PFO, we propose to accelerate the randomized search via updating a \emph{particle swarm template (PST)}, a set of particles which are \emph{pre-sampled} uniformly within the unit sphere in the 6D space of camera pose. Specifically, the PST is \emph{moved and rescaled} progressively, dictated by the so-far best solution found in the particle set which maximizes an observation likelihood. Through evolving the pre-sampled PST, our method is able to drive tens of thousands of particles to locate and cover a good local optimum extremely fast and robustly.

\emph{Secondly}, the sampled particles, representing candidate poses, are evaluated with a depth-based \emph{discriminant fitness/likelihood} function. The fitness of a particle is measured as depth-model conformance through integrating the truncated signed distance field (TSDF) values over the depth map transformed by the corresponding camera pose. This can be efficiently evaluated based on TSDF occupancy queries and requires no frame-to-frame feature correspondence which is difficult under fast motion. Thanks to the efficient template-based particle maintenance and the effective observation likelihood, our method attains good quality pose tracking under fast camera motion (up to $4$m/s) in a realtime framerate ($30$Hz) without depth map filtering, frame dropping, or global pose optimization.

We have evaluated our method on several public benchmarks. On ordinary speed sequences, our method achieves comparable accuracy of camera tracking against the state-of-the-art methods, without needing post-processing such as loop closure detection or global pose optimization. On the \DSETH benchmark~\cite{schops2019bad}, our method successfully reconstructs those challenging sequences with fast camera motion (i.e., shaking cameras) on which all previous methods failed. We also contribute a new benchmark named \DSFCM which encompasses both synthetic and real captured sequences with fast camera motion ($3\sim 4$ times faster than the existing datasets).
Extensive quantitative and qualitative evaluations demonstrate the significant advantage of our method on fast-motion camera tracking and dense reconstruction.

To sum up, the contributions of this work are:
\vspace{-2pt}
\begin{itemize}
	\item {\em Problem:} We study the problem of online dense reconstruction under fast camera motion \kx{without using an IMU} and propose the first solution to it based on random optimization.
	\item {\em Optimization:} We propose a novel particle filter optimization which achieves robustness and effectiveness via evolving a pre-sampled particle swarm template. Through detailed comparisons to the classical particle filter optimization and particle swarm optimization, we show that our method achieves a much better balance between exploration and exploitation.
	\item {\em Benchmark:} We propose the first dataset of fast-camera-motion RGB-D sequences, along with ground-truth trajectories and reconstructions for synthetic and real captured data, respectively.
    \item {\em System:} We have implemented an end-to-end system of online RGB-D dense reconstruction which realizes robust and quality realtime reconstruction under fast camera moving speed up to $4$m/s.
\end{itemize}


\section{Related work}
\label{sec:related}

\paragraph{Online RGB-D reconstruction.}
There is a large body of literature on offline and online RGB-D reconstruction; let us review only those highly related ones. KinectFusion~\cite{Newcombe2011,Izadi2011} is one of the first to realize a real-time volumetric fusion framework of~\citeN{Curless96}.
In order to handle larger environments, spatial hierarchies~\cite{chen2013scalable}, and hashing schemes~\cite{Niessner2013,Kahler2015} have been proposed.
Another line of research adopt point-based representation~\cite{Whelan2012,keller2013real,henry2014rgb,Whelan2015}.

Real-time per-frame camera pose estimation is a core problem in the Simultaneous Localization and Mapping (SLAM) literature. Several real-time monocular RGB methods (e.g.,~\cite{klein2007parallel,engel2013semi,forster2014svo,engel2014lsd}) typically
rely on either pose-graph optimization~\cite{Kuemmerle2011}
or bundle adjustment~\cite{triggs1999bundle}.
In the case of dense reconstruction, MonoFusion~\cite{pradeep2013monofusion} integrates sparse SLAM bundle adjustment with dense volumetric fusion. DTAM~\cite{newcombe2011dtam} estimates camera poses directly from the reconstructed dense 3D model based on frame-to-model tracking.
Based on GPU optimization techniques, several methods~\cite{schops2019bad,Dai2017} realizes real-time global pose alignment.


Common to most existing state-of-the-arts is the reliance on photometric-error-based objective and gradient-descent-based optimization (e.g.~\cite{Dai2017}). Bylow et al.~\shortcite{bylow2013real} realize a feature-free camera tracking via defining an objective function of pose optimization based on depth-to-TSDF agreement. Our fitness function is defined similarly. However, they still employ the gradient descent method in their optimization.
To the best of our knowledge, our work is the first that utilizes random optimization for camera pose estimation, at least in the context of online dense reconstruction.

\kx{
\paragraph{Camera tracking with inertial measurements.}
Visual-inertial solution to camera tracking for SLAM, odometry or online reconstruction is an active field of research~\cite{scaramuzza2019visual} and has been successfully deployed in practice.
IMUs provide acceleration data at a high frequency and can be used to predict inter-frame motions serving as good initialization for gradient-descent-based optimization~\cite{forster2016manifold}.
The gyroscope sensor of IMUs (especially built-in IMUs in commodity RGB-D cameras) is more effective for measuring changes in orientation than estimating translations. Many researchers found the translation error too large to be useful in tracking, either used as pose initialization~\cite{niessner2014combining} or for joint optimization~\cite{laidlow2017dense}.

To our knowledge, most VIO works are mainly designed for large-scale environments with decimeter-level tracking accuracy~\cite{mourikis2007multi,leutenegger2015keyframe}.
The accuracy is primarily determined by visual tracking which is prone to motion blur, although the latter is usually not a significant issue for outdoor environments due to large camera-to-scene distance~\cite{cui2019real,pollefeys2008detailed,engel2015large}.
These existing methods, however, usually cannot support a centimeter-level fast-motion camera tracking in the case of indoor-level camera-to-scene distance.
Saurer et al.~\shortcite{saurer2016sparse} additionally considers the rolling shutter effect under fast camera motion. Another line of research uses event camera for fast motion camera tracking~\cite{gallego2019event}.
}

\paragraph{Pose tracking based on particle filter.}
Let us differentiate our work from the large literature of camera/object pose tracking based on particle filter (PF), and more specifically, Rao-Blackwellized particle filter (RBPF)~\cite{andrieu2002particle}. There has been extensive research on particle filter based SLAM~\cite{grisetti2007improved,gil2010multi,choi2012robust} and object pose tracking~\cite{deng2019poserbpf,nieto2016real,arnaud2005efficient}.

The key difference of the particle filter utility between our method and the line of works above lies in the objective. Taking the camera tracking in SLAM as an example, those previous works (e.g.~\cite{grisettiyz2005improving}) optimize for sequential state (pose) estimation throughout \emph{a sequence of frames} via maximizing observation likelihood. On the other hand, our work optimizes the camera pose of \emph{a single (current) frame}. In short, the sequential importance sampling is across different frames in~\cite{grisettiyz2005improving} and across increasing iteration steps for the pose of a single frame in our work. \Fig{pgm} contrasts the two different problems with their underlying probabilistic graphical models.

\paragraph{Particle Filter Optimization (PFO) and Particle Swarm Optimization (PSO).}
The sequential Monte-Carlo method of particle filter has been employed in solving global optimization problems~\cite{liu2016particle,zhang2017controlled}.
The basic idea of PFO algorithms is to transform the objective function into a target PDF and then perform sequential importance sampling to simulate the target probability density function (PDF). The hope is that the optimum of the objective function can be covered by sampled particles.
PSO is a well studied heuristic optimization technique inspired by the social behavior in nature~\cite{shi1999empirical}. In PSO, a set of particles are generated randomly and moved according to their own experience and the experience of the swarm (swarm intelligence). PSO suffers the premature convergence making it easily get stuck in local optima.

The core of PFO is how to design the system dynamic function which drives the set of particles to move toward good local optima. In a recent attempt, PSO update is used as system dynamic of the state space model~\cite{ji2008particle}. However, sampling and updating particles in PFO is still time-consuming. This explains why PFO has not been widely adopted in realtime applications. Our method improves PFO by moving and rescaling a particle swarm template thus avoiding particle sampling and resampling in standard PF.


\section{Method}
\label{sec:method}

\paragraph{Overview.}
The input to online reconstruction is an RGB-D sequence $\{I_c^t, I_d^t\}_{t=0:T}$ ($I_c$ and $I_d$ are RGB and depth images, respectively) captured by an RGB-D camera and the output is a surface reconstruction of the scene being captured, $\mathcal{S}$, and a trajectory of 6DoF camera poses, $\{[\bR^t|\bt^t]\}_{t=0:T}$ ($\bR \in SO(3)$ and $\bt \in \mathbb{R}^3$ are 3D rotation and 3D translation in the global coordinate system, respectively). We build our method upon the volumetric depth fusion framework~\cite{Curless96,Izadi2011} which is the de facto method for high-quality online dense reconstruction.
The key problem of online reconstruction is the estimation of 6D camera pose for each frame. Based on the per-frame camera pose, the corresponding depth map can be fused into the 3D volume incrementally.
To handle fast camera motion, we rely only on depth input for camera pose tracking, which is our key contribution. Of course, our method could admit RGB image input for camera tracking and relocalization when the camera motion is slow and image features are robustly detected.

This section focuses on the key problem --- depth-based per-frame camera pose optimization. After describing the problem formulation (\Sec{formulation}), we introduce our random optimization framework (\Sec{pfo}). We then introduce how to improve the optimization by replacing the sequential importance sampling with particle swarm template evolution (\Sec{pfopst}).

\subsection{Formulation of Per-Frame Pose Optimization}
\label{sec:formulation}
Given the depth image of the current frame $I_d^t$ and the so-far constructed TSDF $\psi:\mathbb{R}^3\rightarrow\mathbb{R}$, our task is to compute the 6-DoF camera pose of $I_d^t$ in the global coordinate system: $[\bR^t|\bt^t]\in SE(3)$. Hereafter, we would omit the frame index $t$ and use $[\bR|\bt]$ to represent the global camera pose of the current frame to be optimized.

\begin{figure}[t]
\centering
\begin{overpic}
[width=\linewidth]
{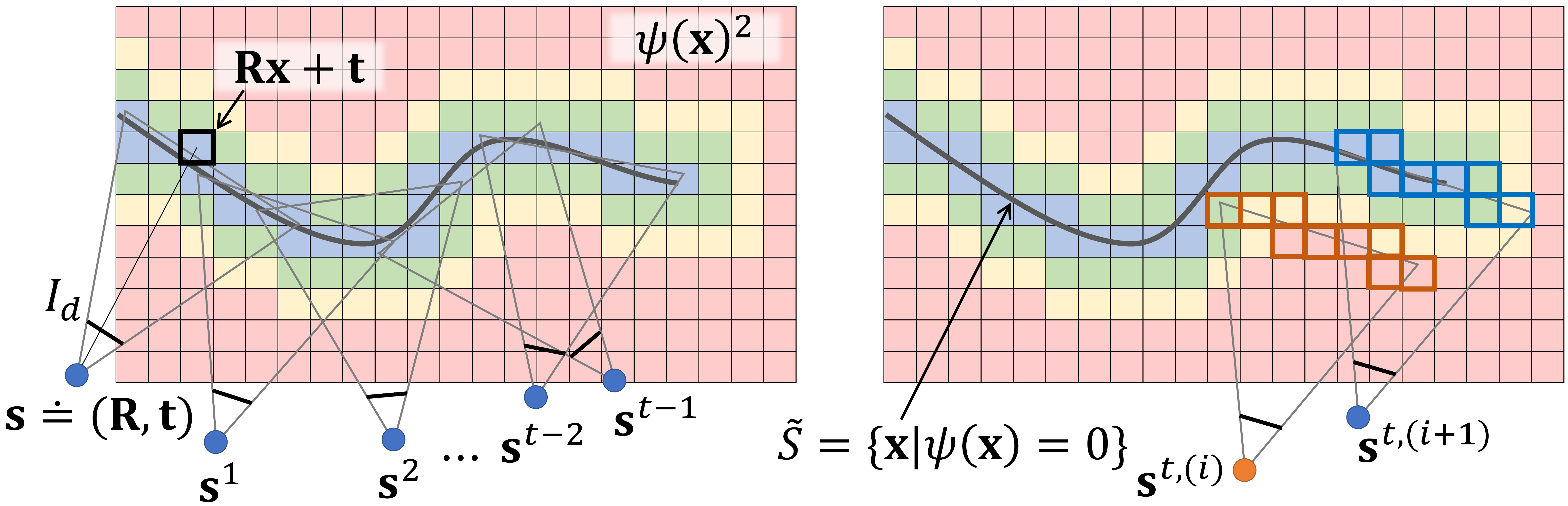}
  \put(28,33.5){\small (a)}
  \put(75,33.5){\small (b)}
\end{overpic}
\caption{
Illustration of per-frame camera pose optimization by minimizing the frame-to-model (depth-to-TSDF) error function in \Eq{nll}. (a): TSDF-based depth fusion and TSDF embedding $\bR\bx+\bt$ of an unprojected 3D point $\bx$ of $I_d$ under a camera pose $(\bR,\bt)$. The grid plot shows squared TSDF values (blue is small and red is large). (b): For frame $t$, the depth map is first embedded into the TSDF volume with the candidate camera poses $(\bs^{t,(i)})_{i=1:N}$. The optimal pose is the one that minimizes the sum of retrieved TSDF values by all unprojected 3D points. In (b), $\bs^{t,(i+1)}$ is a better solution since the summed TSDF values over the blue voxels is lower than that of $\bs^{t,(i)}$ over the orange voxels.
}
\label{fig:tsdfmin}
\end{figure}

We base our approach on the depth-fusion-based pose estimation~\cite{bylow2013real} which defines a frame-to-model error metric to evaluate the goodness of a camera pose $[\bR|\bt]$ by measuring how well $I_d$ ``fits into'' the TSDF under $[\bR|\bt]$.
For each pixel $(i, j)$ of $I_d$, suppose its depth is $z=I_d^t(i,j)$ based on which we can obtain its corresponding 3D point $\bx_{ij}$ in the camera coordinate system of the current frame. We can then transform this point to the global coordinate system:
\begin{equation}\label{eq:unproj}
\bx_{ij}^G = \bR \bx_{ij} + \bt.
\end{equation}
We use these unprojected 3D points to query the TSDF (defined in the global coordinate system) and obtain a point-to-surface distance directly. If the camera pose is correct, it is expected that the point-to-surface distances of every unprojected 3D point should be zero. We therefore seek for the camera pose $[\bR|\bt]$ such that every unprojected 3D point from the depth image lies as close as possible to the zero-crossing surface of the TSDF, i.e., $\tilde{\mathcal{S}}=\{\bx|\psi(\bx)=0\}$. See \Fig{tsdfmin} for an illustration.

Assuming that the depth measurements of the camera contain Gaussian noise and that all pixels are independent and identically distributed, the likelihood of observing a depth image $I_d$ from camera pose $[\bR|\bt]$ is
\begin{equation}\label{eq:likelihood}
  p(I_d|\bR,\bt) \varpropto \prod_{i,j}\exp\left(-\psi(\bR\bx_{ij}+\bt)^2\right).
\end{equation}
Therefore, our goal is to find the optimal camera pose $[\bR^\ast,\bt^\ast]$ that maximizes the likelihood
\begin{equation}\label{eq:objective}
  (\bR^\ast,\bt^\ast) = \arg\max_{\bR,\bt}{p(I_d|\bR,\bt)}.
\end{equation}
In~\cite{bylow2013real}, the maximum likelihood estimation is performed by minimizing the following error function by taking the negative logarithm of the likelihood, i.e.,
\begin{equation}\label{eq:nll}
  (\bR^\ast,\bt^\ast) = \arg\min_{\bR,\bt} \sum_{(i, j)\in I_d}{\psi(\bR\bx_{ij}+\bt)^2}.
\end{equation}
Directly optimizing \Eq{nll} using gradient descent methods (e.g. Gauss-Newton used in~\cite{bylow2013real}) is difficult in the case of fast motion due to the high nonlinearity caused by large rotations.
Moreover, the gradient can be undefined when the depth map is out of the valid scope of the TSDF under large camera transformation. We therefore resort to random optimization based on particle filter which leads to simple, efficient and robust camera tracking under fast motion.
Next, after a brief review of particle filter, we introduce our random optimization solution to pose optimization based on particle filter optimization.

\subsection{Particle Filter Pose Optimization}
\label{sec:pfo}

\paragraph{Particle filter}
(PF) is a class of importance sampling and resampling techniques designed to simulate from a sequence of probability distributions for sequential inference problems~\cite{gordon1993novel}.
It has gained significant success in non-Gaussian and non-linear state estimation problems.

Given a state space model,
\begin{eqnarray}
  & s_k = f(s_{k-1}) + w_k, \\
  & o_k = g(s_k) + v_k,
\end{eqnarray}
where $f(\cdot)$ is the system dynamic, $g(\cdot)$ is the observation function, and $w_k$ and $v_k$ are the system noise and observation noise, respectively. Note that $k$ indicates the time step in sequential sampling; it has nothing to do with the frame index in the context of our work.
We are interested in estimating the posterior distribution $p(s_k|o_{1:k})$, where $o_{1:k}$ denotes the observations obtained until step $k$.
The posterior can be computed in a recursive manner:
\begin{eqnarray}
  & p(s_k|o_{1:k}) \varpropto p(o_k | s_k) p(s_k | o_{1:k-1}), \label{eq:sbi}\\
  & p(s_k|o_{1:k-1}) \varpropto \int{p(s_k|s_{k-1})p(s_{k-1}|o_{1:k-1})dx_{k-1}}, \label{eq:sbi-marginal}
\end{eqnarray}
where $p(s_{k-1}|o_{1:k-1})$ is the posterior at step $k-1$ and $p(o_k | s_k)$ the observation likelihood.
To resolve the intractable integration in \Eq{sbi-marginal}, PF approximates the posterior with a set of weighted Monte Carlo samples or particles $\{s^{(i)}_k, w^{(i)}_k\}_{i=1:N}$
\begin{equation}\label{eq:pf-posterior}
  p(s_k|o_{1:k}) \thickapprox \sum_{i=1}^{N} w^{(i)}_k \delta\left(s_k-s^{(i)}_k\right),
\end{equation}
where $\delta$ is the Dirac delta function.

PF proceeds as follows.
In each step, a set of particles $\{s^{(i)}_{k}\}_{i=1:N}$ is generated from a proposal function:
\begin{equation}\label{eq:proposal}
  s^{(i)}_{k} \thicksim q(s_{k} | s^{(i)}_{k-1}), \quad i=\{1,\ldots,N\}.
\end{equation}
Here, the proposal function $q(s_{k} | s_{k-1})$ usually takes some system dynamic prior. The importance weights of the particles are updated by multiplying the likelihood:
\begin{equation}\label{eq:weight}
   w^{(i)}_{k} = w^{(i)}_{k-1} p(o_{k}|s^{(i)}_{k}), \quad i=\{1,\ldots,N\}.
\end{equation}
To mitigate the particle depletion issue, a resampling step is invoked which resamples particles according to the updated importance weights. After resampling, regions in state space with higher importance weight will be populated with more particles. The weights of the resampled particles are then reset to $\frac{1}{N}$.

\begin{figure}[t]
\centering
\begin{overpic}
[width=\linewidth]
{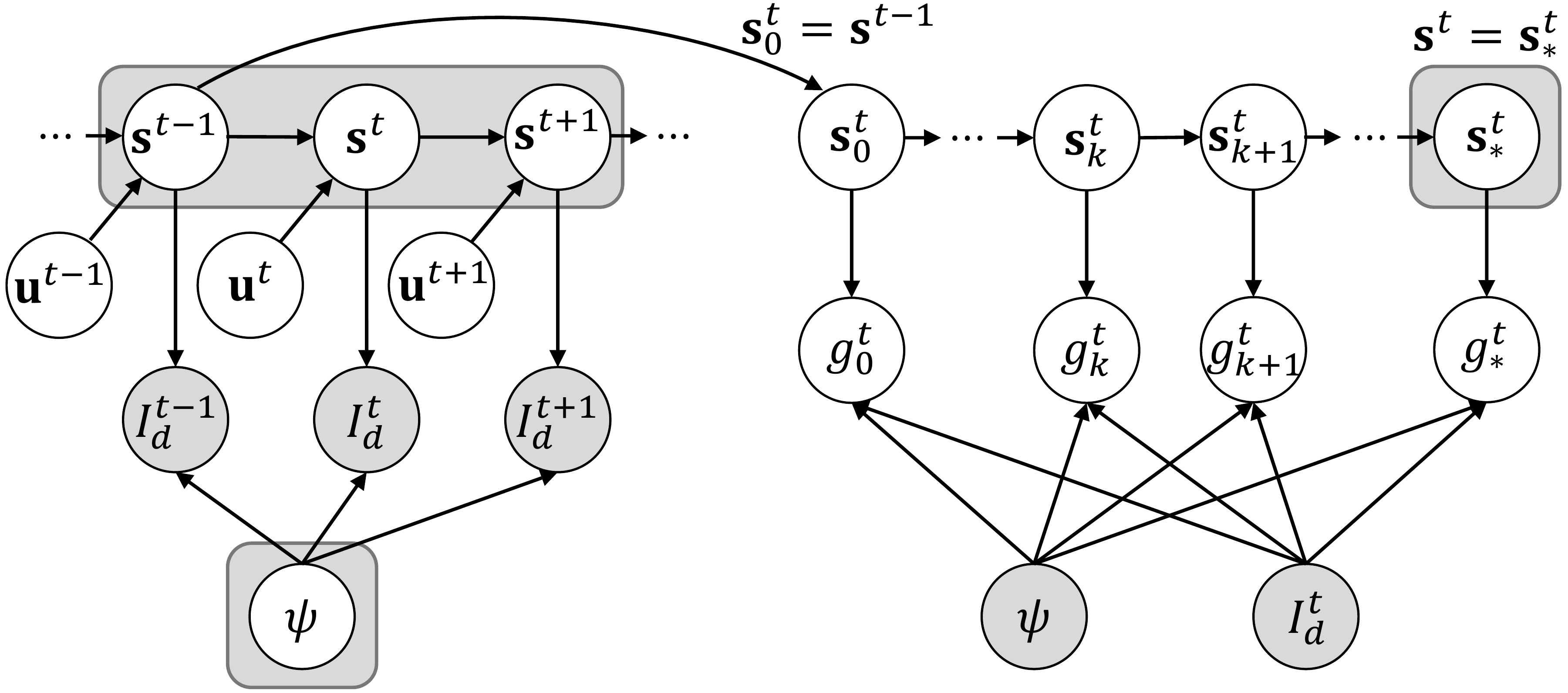}
\end{overpic}
\caption{
The probabilistic graphical models of online reconstruction (left) and per-frame pose optimization (right). Left: The online reconstruction can be represented by a hidden Markov model (HMM)~\cite{thrun2002probabilistic} where the per-frame camera poses $(\bs^t)_{t=0:T}$, the inter-frame camera motions $(\bu_t)_{t=0:T}$ and the TSDF $\psi$ are latent variables, and $(I_d^t)_{t=0:T}$ observed ones. Right: The optimization of $\bs^t$ (camera pose for frame $t$) starts from the initial state $\bs^{t-1}$ and proceeds until the optimal solution $\bs^t_\ast$ is found. Here, we maximize the objective $g^t$ while assuming both the observation $I_d^t$ and the TSDF $\psi$ (after frame $t-1$ is fused) are known. The latent variables enclosed by the rounded rectangles are to be estimated.
}
\label{fig:pgm}
\end{figure}

\paragraph{Particle filter optimization}
(PFO) is a recently proposed random optimization method~\cite{liu2016particle}.
The basic idea is to represent the objective function as a target PDF and then leverage sequential importance sampling to simulate the target PDF. The goal is to cover the optimum of the objective function with samples, i.e., optimizers.
Given a general minimization problem
$$
\min_{s\in S}g(s),
$$
where $S \subset \mathbb{R}^d$ is the non-empty solution space and $g$ a real-valued objective function bounded on $S$. In PFO, the solutions $s$ are treated as states and the objective $g(s)$ as the observation function. PFO performs a sequential sampling of the target posterior distributions $p(s_k|g_{1:k})$ until the optimum state/solution $s^\ast$ is reached.

In each step of PFO, a set of particles $\{s^{(i)}_{k}\}_{i=1:N}$ are sampled from the proposal PDF $q(s_{k} | s_{k-1})$.
A common choice of proposal distribution is symmetric, e.g.,
\begin{equation}\label{eq:proposal}
  q(s_{k}|s^{(i)}_{k-1}) = \mathcal{N}(s^{(i)}_{k-1}, \Sigma),
\end{equation}
where $\mathcal{N}(s^{(i)}_k, \Sigma)$ is a normal distribution with mean $s^{(i)}_k$ and covariance matrix $\Sigma$.
According to \Eq{weight}, particle weights are updated by multiplying the likelihood function defined as follows:
\begin{equation}\label{eq:pfolh}
  p(g|s_{k}) = \exp\left(-\frac{g(s_{k})-g(s^\ast)}{\tau}\right),
\end{equation}
where $\tau$ is a temperature in the Boltzmann distribution (we set $\tau=1$ by default).
Based on the updated weights, particles are resampled and their weights are reset to be uniform.

\paragraph{PFO for camera pose optimization.}
In our problem setting, we treat the camera pose $\bs=(\bR, \bt)$ as state.
Our goal is the minimization problem stated in \Eq{nll}.
The likelihood function corresponding to the objective in \Eq{nll} is then defined as:
\begin{equation}\label{eq:ourlh}
p(g|\bs,I_d) = \exp\left(-\frac{1}{\tau}\sum_{(i, j)\in I_d}{\psi(\bR\bx_{ij}+\bt)^2}\right),
\end{equation}
given that a lower bound estimation of $g(\bs^\ast)$ is $0$.
See the probabilistic graphical model of PFO-based per-frame camera pose optimization in \Fig{pgm} and contrast it with that of online reconstruction. The former is a subproblem of the latter.

\begin{figure}[t]
\centering
\begin{overpic}
[width=\linewidth]
{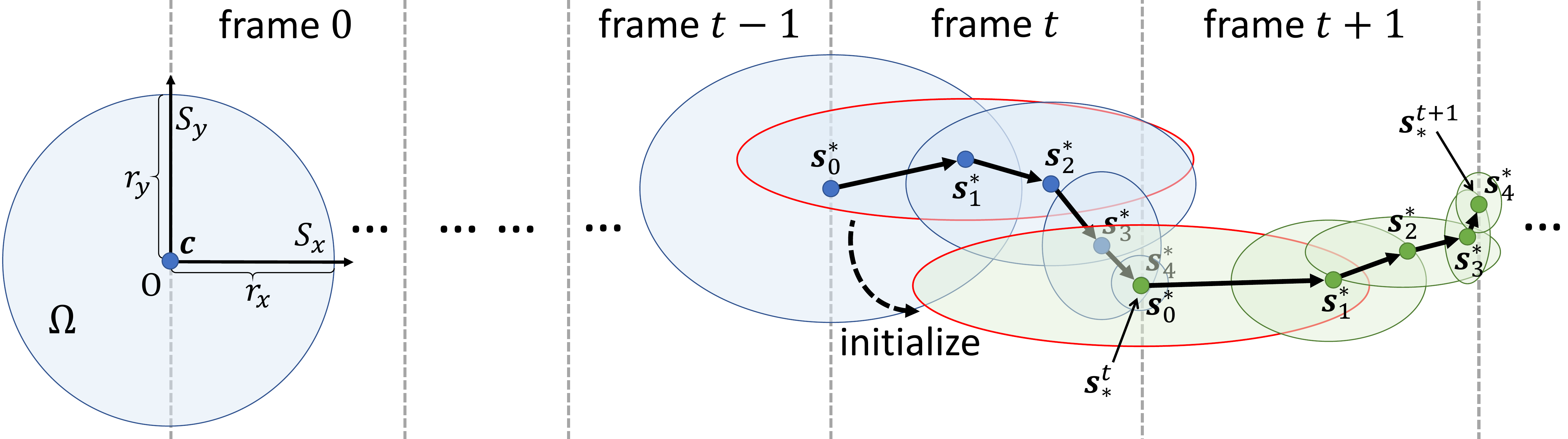}
\end{overpic}
\caption{
Per-frame pose optimization by PFO with evolving PST. Starting with a PST, $\Omega$, sampled from the unit sphere in the parametric space of 6D camera pose (illustrated with 2D here), we optimize the camera pose for each frame through evolving the PST. For each frame $t$, a sequential search is performed by moving and rescaling the PST, leading to a sequence of interim solutions $(\bs_k^\ast)_{k=1:K}$, until a sufficiently good solution (camera pose), $\bs^t_\ast$, is covered by the PST particles. This figure also illustrates PST initialization for a given frame. For example, the PST of frame $t+1$ is initialized at $\bc_0^{t+1}=\bs^t_\ast$, with the axis lengths of the PST after the first iteration of frame $t$, i.e., $\br_0^{t+1}=\br_1^{t}$; see the PSTs highlighted with red boundary.
}
\label{fig:pstevolve}
\end{figure}

A critical issue now is how to select a good proposal distribution $q(\bs_{k}|\bs^{(i)}_{k-1})$ to enable an efficient exploration of the solution space.
A naive system dynamic such as the normal distribution in \Eq{proposal} cannot be very efficient since it does not exploit the information from the previous step. A recent trend on improving sample efficiency of particle filter is to improve the system dynamic with the \emph{swarm intelligence} of all particles~\cite{ji2008particle}. However, sampling and maintaining a particle swarm is computationally expensive. We propose to replace the sequential sampling of PFO by moving and rescaling a pre-sampled particle swarm template.

\subsection{PFO with Particle Swarm Template}
\label{sec:pfopst}
Let us first define the solution/state as the camera pose $\bs=(\bR, \bt)\doteq(q_x, q_y, q_z, x, y, z)$ with $q_x$, $q_y$ and $q_z$ being the imaginary part of the rotation quaternion and $\bt=(x,y,z)^T$.
In our PFO, instead of sampling the particles on-the-fly throughout the optimization, we opt to \emph{pre-sample} a fixed number of particles uniformly within the unit sphere in the 6D state space.
The pre-sampled particle set is referred to as Particle Swarm Template (PST), denoted by $\Omega$.
Let us represent the PST by a center position $\bc$ and a vector of axis lengths (each for one of the six dimensions) $\br=(r_d)_{d=1:6}$.
Initially, PST is sampled from the unit sphere hence $\bc=\bzero \in \mathbb{R}^6$ and $\br=\bone \in \mathbb{R}^6$. It is then moved and rescaled into ellipsoids throughout the optimization steps to gradually cover the optimal solution.

\Algo{pfopst} describes our PFO-based per-frame pose optimization with PST.
The algorithm evolves a pre-sampled PST with moving and rescaling through optimization steps.
The process repeats until a satisfactory solution $\bs^\ast$ is obtained or the maximum search step is reached.
See \Sec{impl} for the details on PST initialization and termination criteria.
\Fig{pstevolve} illustrates the optimization process.

\IncMargin{0.5em}
\begin{algorithm}[t]
\caption{Per-frame Pose Optimization based on PFO with PST}
\label{algo:pfopst}
\SetCommentSty{textsf}
\SetKwInOut{AlgoInput}{Input}
\SetKwInOut{AlgoOutput}{Output}
\SetKwFunction{BestState}{CompBestState}
\SetKwFunction{AxisLen}{CompAxisLength}
\SetKwFunction{SamplePST}{SamplePST}
\SetKwFunction{Move}{MovePST}
\SetKwFunction{Rescale}{RescalePST}
\SetKwFunction{Init}{Initialize}
\AlgoInput{ Depth map: $I_d$; Pre-sampled PST: $\Omega$.}
\AlgoOutput{ A local optimum: $\bs^\ast$. }
$\Omega_0,\bc_0,\br_0 \leftarrow$ \Init{$\Omega$}\tcp*{\small see details in \Sec{impl}} \label{algol:init}
$\bs^\ast_0 \leftarrow \bc_0$\;
$k \leftarrow 1$\;
\Repeat {Stop condition is met} {
    $\Omega_k^a \leftarrow \varnothing$\;
    \ForEach(\tcp*[f]{collect APS}){$ \bs_k^{(i)} \in \Omega_k $}{
        $\rho(\bs_k^{(i)}) \leftarrow p(g|\bs_k^{(i)},I_d)$\tcp*{\small evaluate fitness with \Eq{ourlh}}
        \If {$ \rho(\bs_k^{(i)})>\rho(\bs^\ast_{k-1}) $} {
            $\Omega_k^a \leftarrow \Omega_k^a \cup \{ \bs_k^{(i)} \}$\;
        }
    }
    $\bs_k^\ast$ $\leftarrow$ \BestState{$\Omega_k^a$, $s_{k-1}^\ast$}\tcp*{\small \Eq{ebs}}
    $\bc_k \leftarrow \bs^\ast_{k}$\tcp*{\small move PST center to $\bs^\ast_{k}$}
    $\br_k$ $\leftarrow$ \AxisLen{$\bs_k^\ast$, $\bs_{k-1}^\ast$, $\br_{k-1}$}\tcp*{\small \Eqs{rsdir}{momentum}}
    $\Omega_{k+1} \leftarrow$ \Move{$\Omega_k$, $\bc_k$, $\bc_{k-1}$}\;
    $\Omega_{k+1} \leftarrow$ \Rescale{$\Omega_{k+1}$, $\bc_{k}$, $\br_{k}$, $\br_{k-1}$}\;
    $k \leftarrow k+1$\;
}(\tcp*[f]{\small see details in \Sec{impl}})
\end{algorithm}
\DecMargin{0.5em} 
\begin{figure}[b]
\centering
\begin{overpic}
[width=\linewidth]
{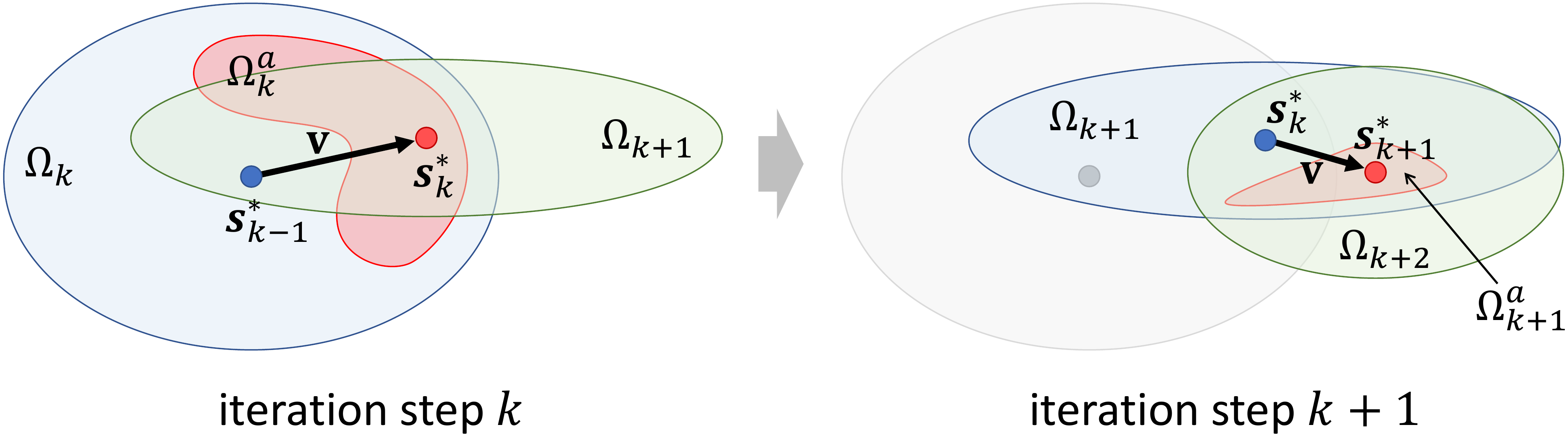}
\end{overpic}
\caption{
Moving and recaling PST from iteration step $k$ to $k+1$. At each step $k$, we first identify the Advantage Particle Set (APS), i.e., $\Omega^a_k$ (shaded in red) which is a subset of the current PST $\Omega_k$ (blue ellipse), and then compute the current best solution $\bs_k^\ast$ (red dot) as the weighted average of the particles in $\Omega^a_k$. The PST $\Omega_k$ is then moved to $\bs_k^\ast$ with the new axis length proportional to the vector $\bv = \bs^\ast_k - \bs^\ast_{k-1}$, thus evolving into $\Omega_{k+1}$ (green ellipse). The same process goes for step $k+1$ shown to the right.
}
\label{fig:pstmove}
\end{figure}

\paragraph{PST move.}
In each iteration step $k$, we first evaluate the fitness $\rho$ for each particle in $\Omega_k$ based on \Eq{ourlh}: $\rho(\bs)=p(g|\bs,I_d)$.
Among all the particles in $\Omega_k$, we select those whose fitness is higher than $\bs^\ast_{k-1}$ (the best solution in step $k-1$) into an \emph{Advantage Particle Set (APS)}: $\Omega^a_k=\{\bs_k^{(i)}\in \Omega_k \mid \rho(\bs_k^{(i)})>\rho(\bs^\ast_{k-1})\}$.
%
With APS, we define the best state at the current step $k$ as the weighted average of all particles in APS:
\begin{equation}\label{eq:ebs}
  \bs^\ast_k = \sum_{\bs_k^{(i)}\in\Omega_k^a}{\bar{\omega}^{(i)}\bs_k^{(i)}}
\end{equation}
where $\bar{\omega}^{(i)}=\omega^{(i)}/\sum_{\bs_k^{(i)}\in \Omega_k^a}{\omega^{(i)}}$ with $\omega^{(i)}=\rho(\bs_k^{(i)})-\rho(\bs^\ast_{k-1})$.

We then move the PST to be centered at the best state of the current step $\bs^\ast_{k}$; see \Fig{pstmove}.
The move in the 6D state space is implemented by the function \FuncSty{MovePST} in \Algo{pfopst}. Note that it is not simply a vector space translation/addition since rotation is not closed under addition. \supl{The details are given in the supplemental material.}

\paragraph{PST rescale.}
Similar to PF, a resampling step is required to avoid particle depletion.
In the context of PST, we instead rescale the spherical or ellipsoidal PST into a new one (\Fig{pstmove}).
In particular, we compute the axis lengths $\br_k$ of the current step as follows:
\begin{eqnarray}
  & \bv = \bs^\ast_k - \bs^\ast_{k-1},\label{eq:rsdir}\\
  & \hat{\br}_k = (1-\rho(\bs^\ast_k)) \frac{\bv}{\|\bv\|} + \bepsilon.\label{eq:rsinter}
\end{eqnarray}
Here, $\bv$ is an anisotropic attractor which drives the particles towards the best solution $\bs^\ast_k$ (the global best of the particle swarm). $\hat{\br}_k$ is the (interim) vector of axis lengths of $\Omega_{k}$, which is scaled by the inverse best fitness $1-\rho(\bs^\ast_k)$. This means that a smaller search range is preferable when the solution is getting better, which helps the optimization converge more stably. $\bepsilon$ is a 6D vector of small numbers ($10^{-3}$) used to avoid degenerating PST.

The final shape of the PST is a blend between the axis lengths $\hat{\br}_k$ and those of the previous step $\br_{k-1}$:
\begin{equation}\label{eq:momentum}
    \br_k = \beta \br_{k-1} + (1-\beta) \hat{\br}_k,
\end{equation}
where we use $\beta=0.1$. This sequential averaging smoothes out the variance of axis lengths across iterations and helps to stabilize the optimization. This is similar in spirit to the momentum mechanism in the Stochastic Gradient Descent (SGD) methods~\cite{ruder2016overview}.
Finally, we compute the scaling factor using $\br_k$ and $\br_{k-1}$ and transform the particles in PST using the function \FuncSty{RescalePST} in \Algo{pfopst}. \supl{See the supplemental material for details.}



\section{Implementation details}
\label{sec:impl}

\begin{figure}[t]
\centering
\begin{overpic}
[width=\linewidth]
{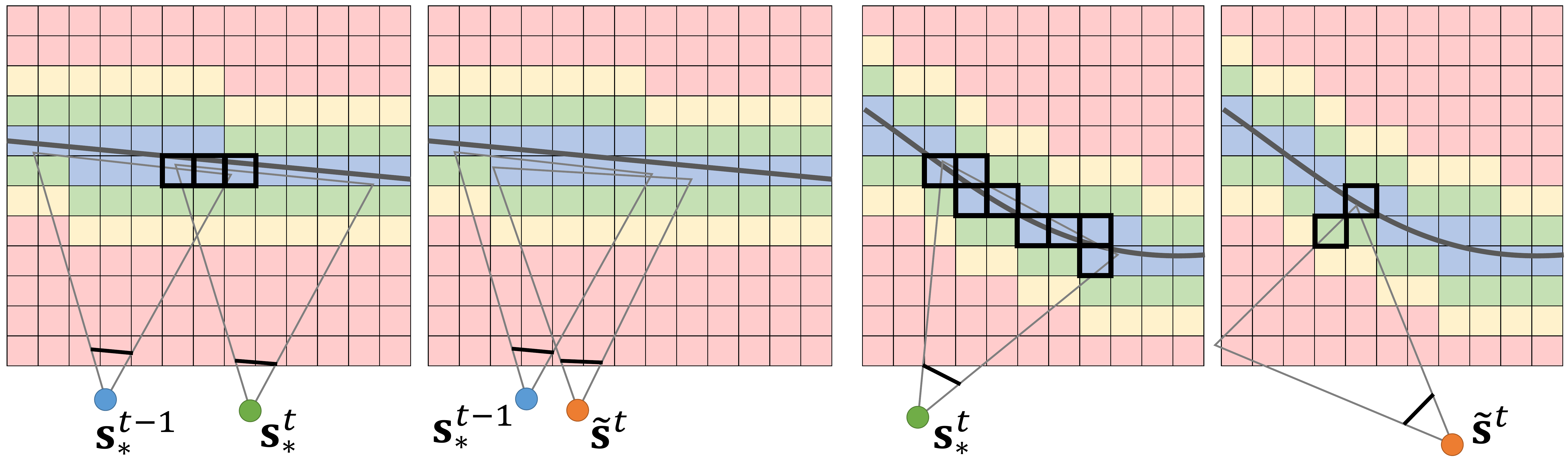}
    \put(12,30.5){\footnotesize (a)}
    \put(38,30.5){\footnotesize (b)}
    \put(65,30.5){\footnotesize (c)}
    \put(87.5,30.5){\footnotesize (d)}
\end{overpic}
\caption{
Finding the correct set of TSDF voxels for likelihood estimation of $\bs^t$. (a-b): When two consecutive frames $t-1$ and $t$ have relatively small overlap (shown as highlighted voxels) due to fast camera motion, using voxels corresponding to all unprojected 3D points of the two depth maps would cause over-alignment of the two frames and a suboptimal solution $\tilde{\bs}^t$ (b). (c-d): If we select only those TSDF voxels with valid values (the highlighted voxels within non-truncated area), a suboptimal solution (d) would have a falsely high likelihood since only two valid voxels are counted.
}
\label{fig:overlap}
\end{figure}
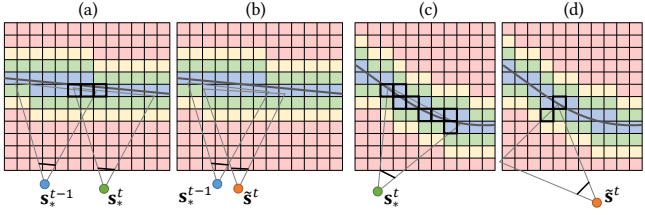

\paragraph{TSDF normalization in likelihood estimation.}
In optimizing the camera pose for a given depth frame, we have devised a correspondence-free approach with TSDF-based maximum likelihood estimation. Essentially, it minimizes the TSDF queries of a depth map under the camera pose. The likelihood is estimated with \Eq{ourlh} which involves the summation of the TSDF values over the 3D points unprojected from \emph{all} depth pixels.
Generally, when performing pose estimation with frame-to-frame registration, one should align \emph{only} the overlapping area observed by both the current and the previous frames;
trying to align the entire depth map as we do with \Eq{ourlh} may cause severe mismatching; see \Fig{overlap}(a,b) for an illustration.

In order to make the likelihood truly reflects how well the current frame aligns with the previous one, we need to determine the overlapping area between the two frames \kx{(note it is not frame-to-TSDF overlap which would cause the likelihood over-evaluates misalignment)}. One quick option is to sum over only those pixels whose unprojected 3D point lies in a valid (updated before) TSDF region. This may also lead to suboptimal solutions as illustrated in \Fig{overlap}(c,d).
To identify the overlapping pixel set $O^t$ of depth frame $I^t_d$ against $I^{t-1}_d$, we adopt the unproject-and-reproject scheme:
\begin{equation}\label{eq:validness}
  O^t = \{(i,j) \mid \pi^{t-1}(\pi^{t})^{-1}[(i,j)] \in I^{t-1}_d \textmd{ with } (i,j) \in I^t_d\},
\end{equation}
where $\pi^{t}$ is the projection matrix of frame $t$ under camera pose $\bs^t$.
In our implementation, the selection of valid pixel set is done for each iteration during the optimization of $\bs^t$. In particular, when computing $O^t_k$ for iteration step $k$, we use the projection $\pi_{k}^{t}$ built with $\bs^{t\ast}_{k-1}$ which is the best pose of step $k-1$. The projection $\pi^{t-1}$ simply takes $\bs^{t-1}_\ast$ which is the final optimal pose for frame $t-1$.

Based on $O^t_k$, the likelihood is estimated by summing over the overlapping depth pixels only and normalizing the summation by the number of those depth pixels:
\begin{equation}\label{eq:normallh}
  p(g|\bs^t_k,I^t_d) = \exp\left(-\frac{\sum_{(i, j)\in O^t_k}{\psi(\bR^t_k\bx_{ij}+\bt_k^t)^2}}{\tau|O^t_k|}\right).
\end{equation}
Note that the inter-frame overlap is generally not too small so that no over-evaluation would happen for those poses leading to small overlap: Our PST scaling scheme ensures that the sampled transformation is usually no greater than $10$cm in translation and $10^\circ$ in rotation w.r.t. the pose of the previous step.

\paragraph{PST pre-sampling.}
Since we sample particles for PST within a unit sphere in the 6D state space, it is guaranteed that any sampled particle represents a valid rigid body motion in $SE(3)$. The translation is measured in meter, hence each optimization step could explore a maximum range of $1$ meter for translation and $2\pi$ for rotation. To realize an unbiased particle sampling, we adopt the 6D Poisson disk sampling proposed in~\cite{bridson2007fast}.

\paragraph{PST initialization.}
For the first frame, PST is initialized with $\bc_0$ and $\br_0$.
For each following frame $t$, we initialize PST with the final PST center of the previous frame $t-1$, i.e., $\bc_0^t=\bs_\ast^{t-1}$, and with the axis lengths after the first iteration of the previous frame, i.e., $\br_0^t=\br_1^{t-1}$; see \Fig{pstevolve}. The rationale of such initialization is as follows. Since the actual solution of the current frame is unknown a priori, a good prior of exploration region is around the best solution of the previous frame. Setting the PST size to be $\br_1^{t-1}$, instead of $\br_0$ or $\br_\ast^{t-1}$, inherits the particle distribution of the previous frame while ensuring a sufficiently large initial search range ($\br_0$ is uninformative and $\br_\ast^{t-1}$ is too small).
In \Algo{pfopst}, the initialization is not written down but instead summarized as a function in \AlgoL{init} since we have omitted frame index in the description.

\paragraph{PST update.}
The computation of the scaling factor in \Eq{rsinter} depends on the evaluation of the likelihood of the current best solution, i.e., $\rho(\bs_k^\ast)$. However, there are two corner cases in which the evaluation cannot be conducted.
\emph{First,}
when TSDF has not been built in the beginning of the reconstruction, the objective likelihood in \Eq{ourlh} cannot be computed. In this case, we use an initial likelihood of $0.5$. The attractor vector in \Eq{rsdir} is set to be isotropic $\bv=\bone\in\mathbb{R}^6$. Hence, the PST becomes a sphere with $\hat{\br}_k=0.5 \bv$.
\emph{Second,}
when a local optimum is reached in the previous iteration step, the current APS is empty and hence $\rho(\bs_k^\ast)$ cannot be computed. If we would like to search for a better solution, we again turn the PST into a sphere by setting $\bv=\bone\in\mathbb{R}^6$ and compute $\hat{\br}_k=2(1-\rho(\bs_{k-1}^\ast))\frac{\bv}{\|\bv\|}$.

\begin{figure}[t]
\centering
\begin{overpic}
[width=\linewidth]
{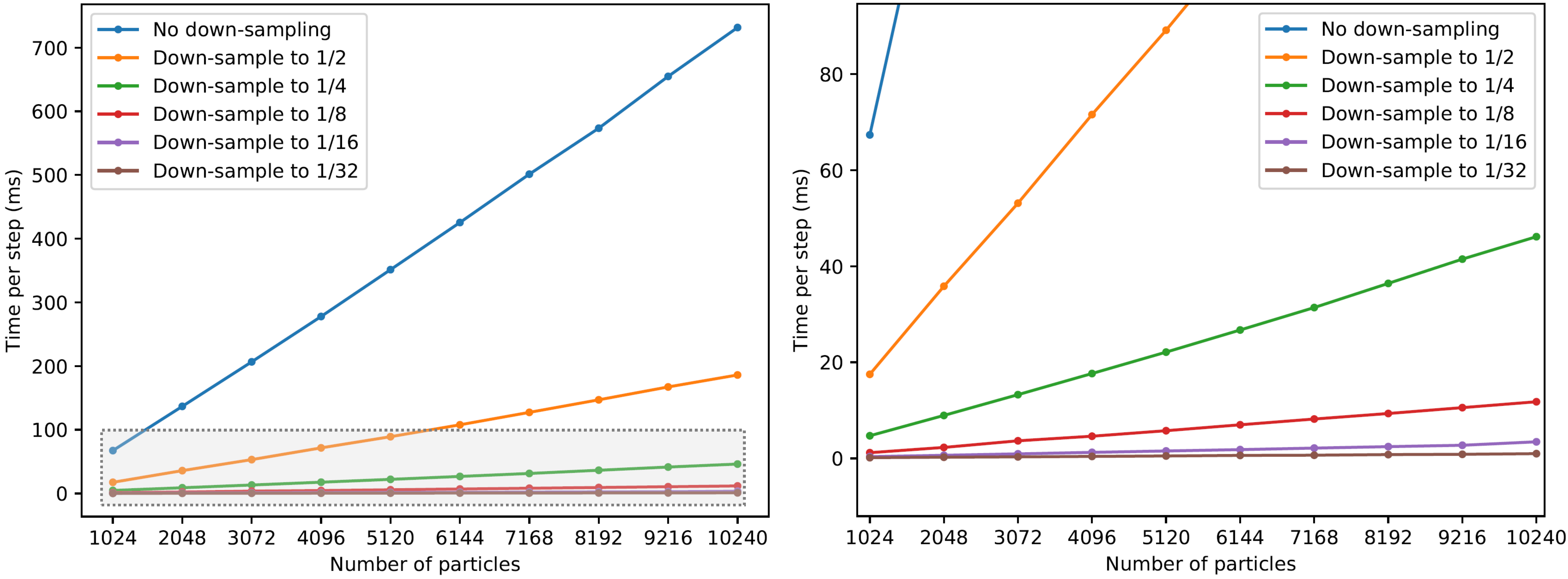}
\end{overpic}
\caption{
The computational time per iteration step for different PST resolutions (number of particles) and different down-sampling rates of depth image. The right plot is a zoom-in view of the grey box in the left plot. The time complexity scales linearly with particle count while down-sampling can significantly reduce the computational cost.
}
\label{fig:plot-complexity}
\end{figure}

\paragraph{Multi-resolution PST.}
The computational cost of our method is proportional to particle count in PST (or PST resolution). While higher PST resolution makes each iteration more time-consuming, it does increase the chance of finding a better local optimal. In order to work with a denser PST, we opt to perform \kx{stride-based downsampling of the depth map} so that the fitness evaluation of each particle could be accelerated. \Fig{plot-complexity} plots the time complexity for different PST resolutions and depth image down-sampling rates. Low-res depth map is, however, more noise sensitive. To balance between the resolutions of PST and depth map, we pre-define the following three combinations of PST resolution and depth map down-sampling rate: ($1024$, $1/8$), ($3072$, $1/16$), and ($10240$, $1/32$). The three combinations are used in turn across different iterations. In \Sec{ablation}, we show that such alternating resolution scheme leads to comparable performance to high-res PST plus full-res depth map with only $1/10$ time consumption.

\paragraph{GPU implementation.}
Following most volumetric RGB-D fusion framework, our method is implemented with the GPU.
Since the fitness evaluation of the particles in PST is time-consuming, we store both TSDF and PST in the GPU memory to facilitate fast computation. To further accelerate the computation, we store the original PST corresponding to the unit sphere and move and rescale this original PST in each iteration with properly computed transformations on the fly. This way, we avoid frequent sampling and resampling of PST particles which involves heavy write operations. All particles are transformed and evaluated in parallel.

\begin{figure}[t]
\centering
\begin{overpic}
[width=\linewidth]
{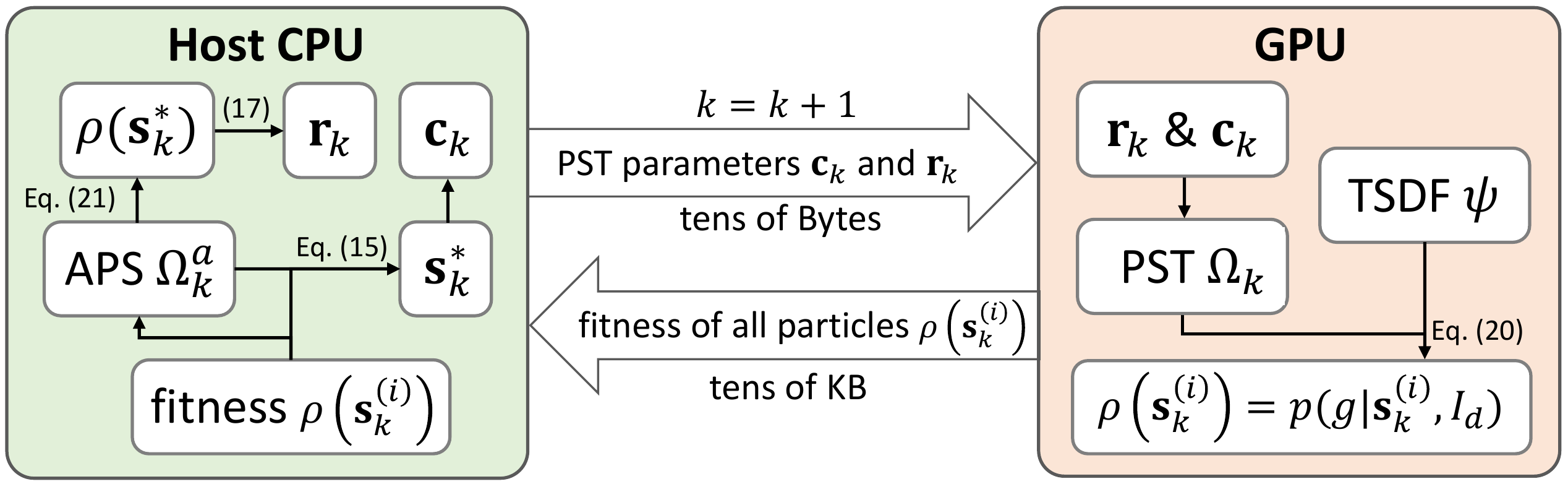}
\end{overpic}
\caption{
The computation tasks of and data streaming between the CPU and the GPU. The evaluation of particle fitness, which is the most time-consuming part, is conducted on the GPU, while the update of PST is done on the CPU. The data flow between is minimal (at most tens of KB).
}
\label{fig:gpu}
\end{figure}

The fitness values of all particles are then streamed into the host CPU. They are used to update the APS, the best solution $\bs_k^\ast$ and the center $\bc_k$ and axis lengths $\br_k$ of PST for the current iteration.
Since the computation involves summation of a large number of floating-point numbers, we employ the Kahan summation algorithm~\cite{neumaier74} which can significantly reduce error accumulation.
The information is then streamed back to the GPU for the next round of PST update and evaluation. With this implementation, the amount of data streaming is minimal (up to tens of KB), thus maximizing the utility of both GPU and CPU. Note, however, the computation of $\br_k$ requires the fitness evaluation of $\bs_k^\ast$ which calls for a GPU computation. To save this extra cost, we instead approximate its fitness by the weighted average of those in the corresponding APS:
\begin{equation}\label{eq:wafit}
  \tilde{\rho}(\bs^\ast_k) = \sum_{\bs^{(i)}_k\in\Omega^a_k}{\bar{\omega}^{(i)}\rho(\bs^{(i)}_k)}.
\end{equation}
The weights are the same as those in \Eq{ebs}. Both the fitness and the weights of APS particles have already been computed by the GPU in the previous round. See \Fig{gpu} for a summary of the computation and data flow on the CPU and GPU.


\paragraph{Convergence criteria.}
We set the maximum iteration step to $20$ to make sure a real-time frame rate (30fps). When we set the termination condition as ``APS is empty for two consecutive iteration steps'', our optimization usually converges with less than $4$ iterations for ordinary motion ($<1$m/s) and less than $10$ iterations for fast motion ($2\sim4$m/s).
Another stop condition is the change of $\bs_k^\ast$ is less than $1\times 10^{-6}$ for each of its six dimensions.


\section{Results and applications}
\label{sec:result}

\subsection{Benchmark}
\label{sec:benchmark}

\paragraph{Public datasets.}
We evaluate our method on three publicly accessible datasets including \DSICL~\cite{handa2014benchmark}, \DSTUM~\cite{sturm2012benchmark} and \DSETH~\cite{schops2019bad}. \DSICL is a synthetic dataset comprising 4 rendered (noise added) RGB-D sequences. For \DSTUM, four real captured sequences are most often compared in the literature. Both have ordinary camera moving speed (typically slower than 1m/s). For \DSETH, we are especially interested in the three challenging sequences. The three sequences, prefixed with ``camera shake'', mainly contain shaking motions of RGB-D camera and exhibit increasing moving speed from \texttt{camera\_shake\_1} to \texttt{camera\_shake\_3}. Let us refer to these three sequences as \DSETHCS. All these datasets possess ground-truth camera trajectories.

\paragraph{The FastCaMo dataset.}
Given that the existing datasets contain very limited number of fast-motion sequences, we contribute a novel and challenging dataset of RGB-D sequences with fast camera motion, named as \DSFCM. The dataset is composed of a synthetic part (\DSFCMS) and a real captured (\DSFCMR) part.

\DSFCMS is built upon the \DSREPL dataset~\cite{straub2019replica} which collects $18$ highly photo-realistic 3D indoor scene reconstructions. Each scene is represented by a dense mesh with high-resolution textures; some of them contain planar mirror and glass reflectors. We select $10$ room-scale scenes out of the collection. For each scene, we create a camera trajectory through manually selecting a serial of key-frame poses and chaining them into a smooth motion trajectory using pose interpolation similar to~\cite{choi2015robust}. Along the trajectory, we generate an RGB-D sequence by sampling frames according to camera moving speed and rendering RGB and depth images based on the camera poses. We control the linear speed of the camera to vary in $[1, 4]$ m/s and the angular speed in $[0.9, 2.2]$ rad/s. For RGB image, we utilize the renderer provided by the dataset~\cite{straub2019replica}. Depth maps are rendered using the dense mesh. We also synthesize motion blur effect for each RGB image and depth noise for each depth map using the same method as in~\cite{handa2014benchmark}.

\DSFCMR contains $24$ real captured RGB-D sequences with fast camera motion (see speed stats in \Tab{speed-info}) for $12$ scenes (each captured twice). The sequences were captured using Azure Kinect DK. As for ground-truth, we opt not to provide camera trajectories since fast moving camera is quite difficult to track using the visual tracking systems as in \DSTUM and \DSETH. We therefore offer a full and dense reconstruction scanned using the high-end FARO Focus 3D Laser Scanner, serving as ground-truth. We can then evaluate RGB-D reconstruction by comparing the reconstructed surface against the ground-truth (see evaluation metrics below).

\Tab{speed-info} provides the speed information of all datasets being tested. Note that we are unable to provide the actual camera moving speed for \DSFCMR since the ground-truth camera trajectories for those sequences are unknown. Here, we give the approximated velocities estimated based on the trajectories tracked by our method, which merely serve as a reference.

\begin{table}[!t]\centering
\caption{
Statistics on camera moving speed (average linear velocity $\bar{v}$, maximum linear velocity $v_\text{max}$, average angular velocity $\bar{\omega}$ and maximum angular velocity $\omega_\text{max}$) for different benchmark datasets. Note ($\ast$): The speed information for \DSFCMR is approximated with our tracking results.
}\vspace{-5pt}
\scalebox{0.95}{
\setlength{\tabcolsep}{1.8mm}{
\begin{tabular}{l|c|c|c|c}
\hline
Item & $\bar{v}$ & $v_\text{max}$ & $\bar{\omega}$ & $\omega_\text{max}$  \\ \hline\hline
\DSICL   & $0.19$m/s   & $1.84$m/s   & $0.24$rad/s  & $0.97$rad/s \\ \hline
\DSTUM   & $0.24$m/s   & $0.96$m/s   & $0.20$rad/s  & $3.37$rad/s \\ \hline
\DSETHCS & $0.38$m/s   & $0.92$m/s   & $1.94$rad/s  & $5.83$rad/s \\ \hline
\DSFCMS  & $1.68$m/s   & $3.93$m/s   & $0.95$rad/s  & $2.18$rad/s \\ \hline
\DSFCMR$\ast$  & $1.03$m/s   & $4.57$m/s   & $0.93$rad/s  & $5.73$rad/s \\ \hline
\end{tabular}
}}
\label{tab:speed-info}
\end{table} 

\paragraph{Evaluation metrics.}
When ground-truth trajectory is available, we measure the camera tracking quality based on the Absolute Trajectory Error (ATE) following most existing works~\cite{sturm2012benchmark}. To estimate ATE, the trajectory to be evaluated is first rigidly aligned to the ground-truth. ATE is then estimated as the mean of pose differences of all frames.

Besides ATE, we also measure the per-frame pose accuracy based on Translation Error (TE)~\cite{sturm2012benchmark}. Given the ground-truth pose $\bT^t_E$ and the estimated pose $\bT^t_G$ of a frame $t$, TE is computed as:
\begin{equation}\label{eq:te}
\text{TE}=\|trans((\bT_G^t)^{-1}\bT^t_E)\|_2,
\end{equation}
where $trans(\bT)$ takes the translation part of the transformation $\bT$. Note that this metric does not require a pre-alignment of the estimated and ground-truth trajectories. As long as the trajectories start from the same initial pose of the very first frame, Translational Error can always be estimated for the following frames in the reference system of the first frame.

When ground-truth trajectories are unavailable, we measure the reconstruction quality based on ground-truth surface reconstruction. Following~\cite{dong2019multi}, we measure reconstruction completeness and accuracy. The reconstruction completeness is measured as the percentage of inlier portion of the ground-truth surface. The reconstruction accuracy is evaluated by RMS error of all reconstructed points against the ground-truth surface. \supl{Please refer to the supplemental material for detail definition of the two metrics.}

\subsection{Ablation Studies}
\label{sec:ablation}
We conduct a series of ablation studies to verify the necessity of several key design choices in our method. They include the random optimization scheme (PFO with PST), the anisotropic PST rescaling mechanism, the TSDF normalization in likelihood/fitness estimation, and the multi-resolution PST alternation strategy.

\paragraph{Optimization strategy (PFO with PST) -- \Algo{pfopst}.}
The key to the success of our method in handing fast camera motion tracking lies in the carefully designed random optimization scheme, i.e., PFO with PST. To verify this core algorithmic design, we compare our full method with \emph{vanilla PFO (no particle swarm guidance)} and \emph{particle swarm optimization (PSO; no particle swarm template)}. To make a fair comparison, all methods use the same likelihood/fitness function and the same amount of particles.

\begin{figure}[t]
\centering
\begin{overpic}
[width=\linewidth]{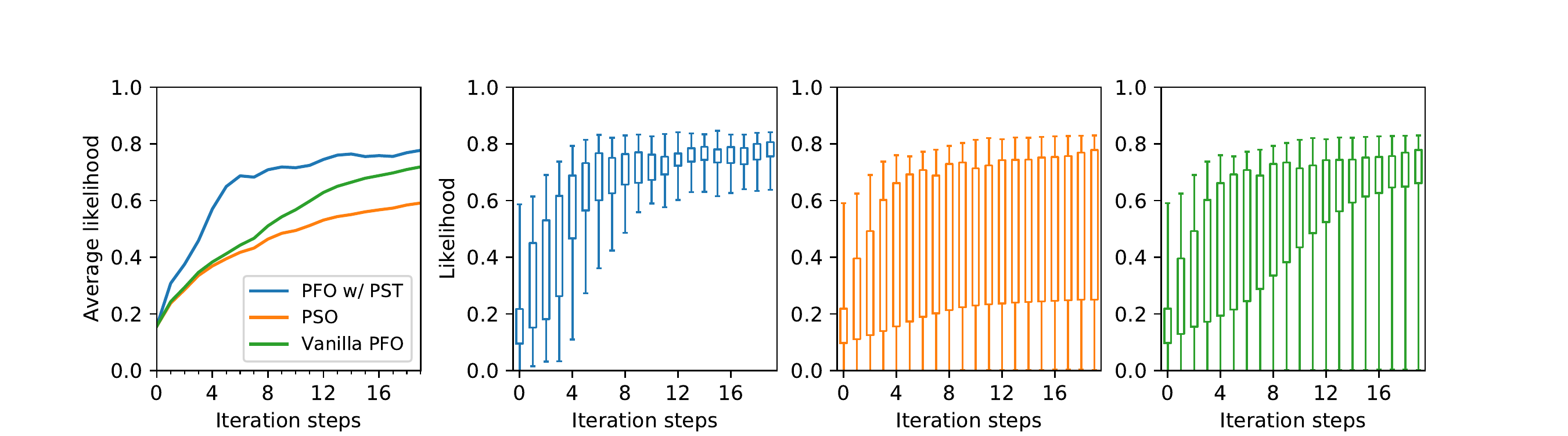}
\end{overpic}
\caption{
Plots of average likelihood/fitness (left column) at different iteration steps for vanilla PFO, PSO and PFO with PST (ours). The right three columns show the range of likelihood over all frames for the three methods.
}
\label{fig:plot-ablt-pfo}
\end{figure}

We evaluate the three methods on $10$ sequences of \DSFCMS. Each sequence contains 6D camera motion in fast speed. In \Fig{plot-ablt-pfo} (left column), we plot for the three methods the average pose likelihood/fitness (estimated using \Eq{normallh}) of all frames and all sequences at different iteration steps. The plot shows that our optimization method leads to the fastest convergence.

In the right three columns of \Fig{plot-ablt-pfo}, we plot at each iteration step the range of pose likelihood values of all frames for the three methods. The range of the medium half of the likelihood values is also highlighted with a slim box. The range of pose likelihood reveals the distribution of the particle fitness and hence the overall uncertainty of the state estimation (solution search). The smaller the range is, the more confidence the particle set is. For PFO-based methods including ours, this means the entropy (diversity) of posterior (or belief) is lower. When a good optimum is approached by the particle set, concentrated (low entropy) is preferred than diverse (high entropy) since the former implies a better particle utilization. Otherwise, a diverse particle set helps to escape from local minima.

From the plots, it can be observed that our method, starting with a diverse set of particles, gradually converges to a low entropy particle set towards a good local optimum, thanks to the PST rescaling mechanism. Without such a smart rescaling mechanism, the vanilla PFO usually gets stuck at a suboptimal solution when particle utilization becomes lower. In PSO, an annealing effect can indeed be observed. However, it tends to drive all particles toward a common point once a good solution is found at that point, which may hurt the exploration ability of the swarm. Our method, on the other hand, achieves a good balance between exploration and exploitation.

\begin{figure}[t]
\centering
\begin{overpic}
[width=\linewidth]
{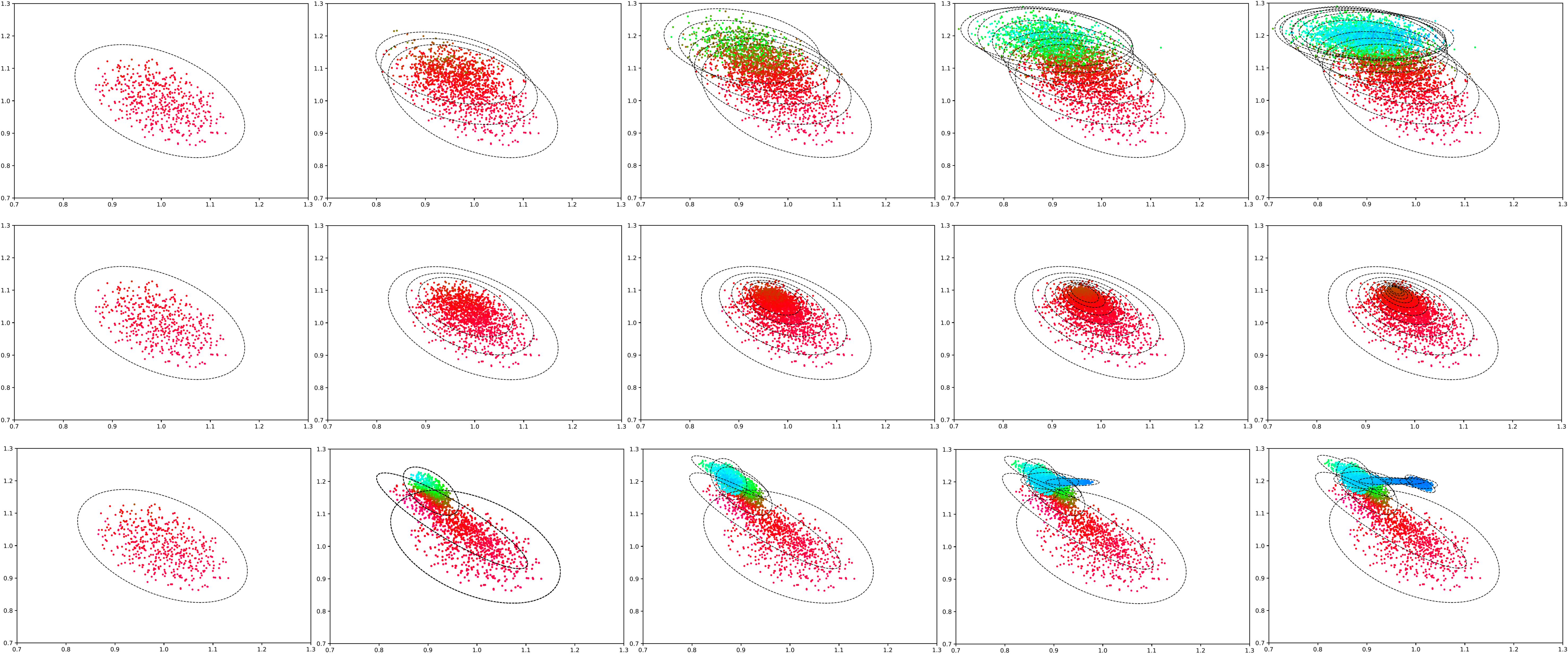}
\put(-3,28.5){\rotatebox{90}{\footnotesize Vanilla PFO}}
\put(-3,19){\rotatebox{90}{\footnotesize PSO}}
\put(-3,-1){\rotatebox{90}{\footnotesize PFO w/ PST}}
\end{overpic}
\caption{
2D visualization of the 6D pose optimization process for a given frame under fast camera motion.
For the three methods, we plot the evolution of particles (dots) over iteration steps where each ellipse indicates the particle set at a specific step and the colors encode the objective function (\Eq{nll}) values (the lower, i.e. the closer to blue, the better) of particles. PFO with PST achieves a better trade-off between exploration and exploitation.
}\vspace{-10pt}
\label{fig:plot-2d-opt}
\end{figure}

\Fig{plot-2d-opt} demonstrates a 2D visualization of the process of pose optimization by the three methods. The plots show the progressive evolution of particle sets over iteration steps. The 6D particle sets are embedded in 2D using sparse random projection~\cite{li2006very}. Starting with the same particle set, the three methods exhibit different exploration behaviors. PFO is able to explore a relatively large range but the particle utilization is low. PSO quickly gets stuck in a local optimum when initialized far from the optima. Our method addresses both issues and results in a much better local optimum.

\paragraph{Anisotropic PST rescaling -- \Sec{pfopst} -- \Eq{rsinter}.}
A core design of our PFO is the anisotropic rescaling of PST guided by particle swarm intelligence. It helps redistribute/propagate the particles smartly so that a better optimum can be covered by the particles more efficiently. Recall that in \Eq{rsinter} the computation of the 6D rescaling factor is comprised of an anisotropic part which is the anisotropic attractor $\bv$, and an isotropic part which is the inverse of likelihood/fitness of the current best solution: $1-\rho(\bs^\ast_k)$. In this study, we evaluate the effect of these two parts individually.

To reveal the effect of anisotropic rescaling in various degrees of freedom, we create three sets of sequences with fast camera motion based on \DSFCMS. The first set contains $10$ sequences where the camera moves with only 1D transformation. In particular, the camera either translates along or rotates about one of the three canonical axes, with a randomly varying linear velocity in $[0.5,4]$ m/s and angular velocity in $[0.2,5]$ rad/s. The second set includes another $10$ sequences of random 3D rotation (yaw, pitch and roll) with the angular velocity ranging in $[0.2,5]$ rad/s. The third set of $10$ sequences are the pre-generated ones with fast 6D camera motion.

\begin{figure}[t]
\centering
\begin{overpic}
[width=\linewidth]{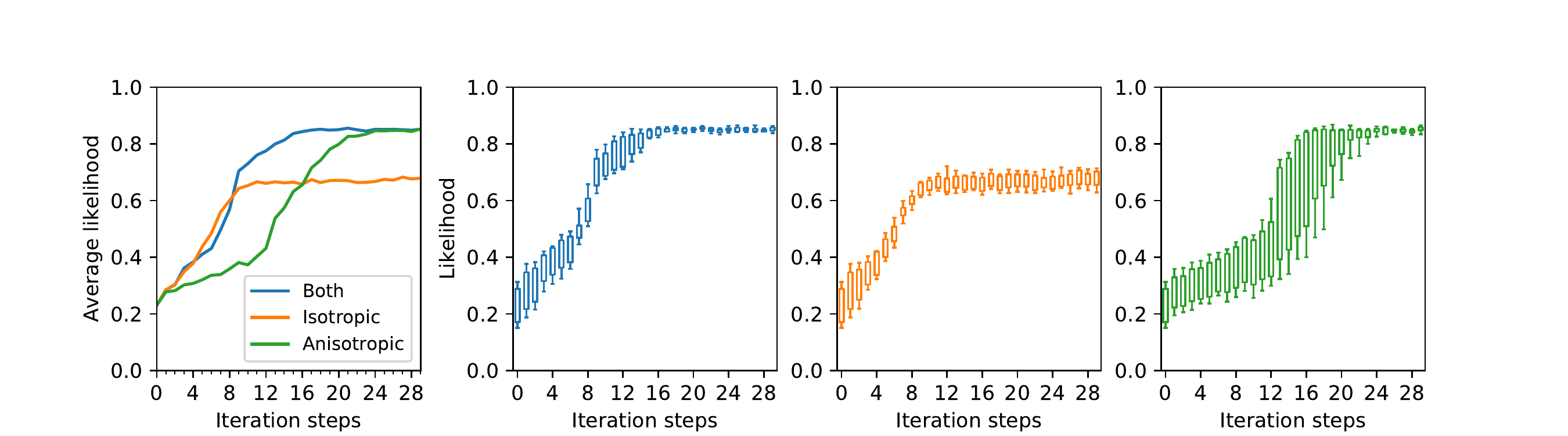}
\put(-2,4){\rotatebox{90}{\footnotesize 1D transformation}}
\end{overpic}
\begin{overpic}
[width=\linewidth]{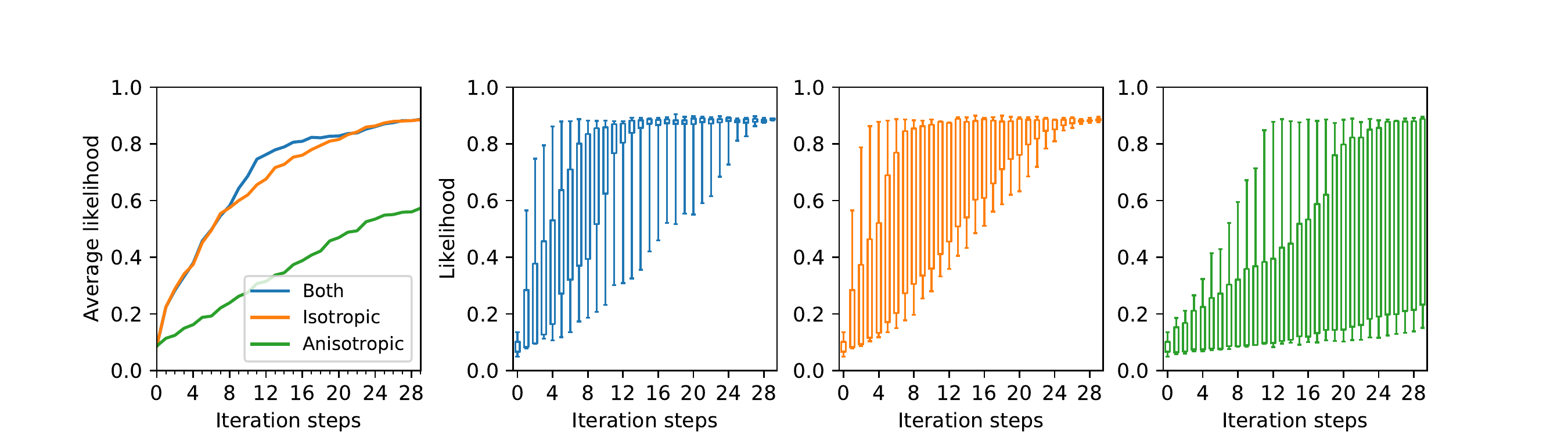}
\put(-3,8){\rotatebox{90}{\footnotesize 3D rotation}}
\end{overpic}
\begin{overpic}
[width=\linewidth]{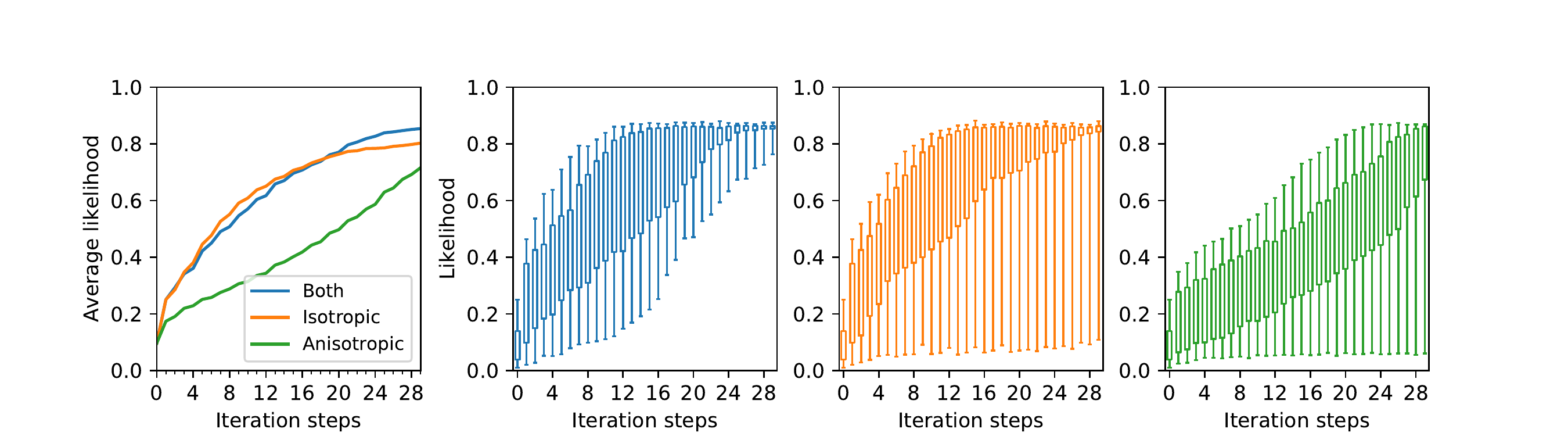}
\put(-3,4){\rotatebox{90}{\footnotesize 6D transformation}}
\end{overpic}
\caption{
Plots of average likelihood/fitness (left column) at different iteration steps for PST update with anisotropic rescaling only, with isotropic rescaling only and with both. The right three columns show the range of likelihood/fitness over all frames for the three methods, respectively. The results are reported for the three datasets of 1D transformation (top row), 3D rotation (middle row) and 6D transformation (bottom row).
}
\label{fig:plot-ablt-rescale}
\end{figure}

For each of the three sets (row), we plot in \Fig{plot-ablt-rescale} (left column) the average pose likelihood/fitness (see \Eq{normallh}) of all frames and all sequences at different iteration steps, using methods \emph{with anisotropic part only}, \emph{with isotropic part only} and \emph{with both}.
The plots show that our PST recaling with both isotropic and anisotropic parts leads to the fastest convergence of optimization (maximization of likelihood/fitness).

In the right three columns of \Fig{plot-ablt-rescale}, we plot at each iteration step the range of likelihood values of all frames for the three methods. For the 1D motion sequence, anisotropic rescaling leads to a more directional distribution of particles and thus contributes more to fast convergence. Isotropic rescaling helps concentrate particles around good optima thus leading to a less scattered likelihoods. Our full rescaling scheme integrates the advantages of the two. For sequences with 3D or 6D motions, however, the difference between isotropic and anisotropic is less obvious due to the entangling of different DoFs. Nevertheless, our full scheme still attains both faster convergence with higher confidence.

\begin{table}[!t]\centering
\caption{
Ablation study on TSDF normalization via evaluating camera tracking accuracy (ATE) on 4 ordinary sequences of \DSICL and 10 fast-motion ones of \DSFCM (the bottom 10 rows). The best results for each sequence are highlighted in \best{blue} color. All the three factors are indispensable while ``No Overlap'' is the most critical one.
}\vspace{-5pt}
\scalebox{0.95}{
\setlength{\tabcolsep}{1.0mm}{
\begin{tabular}{l|c|c|c|c}
\hline
Method               & No Overlap & No Normalize & No Kahan  & Ours \\ \hline\hline
\DSICL \texttt{kt0}    & $4.5$cm   & $3.7$cm   & $0.9$cm  & \best{0.8}cm \\ \hline
\DSICL \texttt{kt1}    & $12.2$cm   & $0.7$cm & $0.7$cm & \best{0.7}cm \\ \hline
\DSICL \texttt{kt2}    & $1.4$cm  & $1.1$cm  &  $1.0$cm & \best{1.0}cm \\ \hline
\DSICL \texttt{kt3}    & $56.0$cm   & $5.4$cm & $9.7$cm & \best{4.5}cm \\ \hline
\texttt{Apartment\_1}    & $2.8$cm   & $1.7$cm & $1.1$cm & \best{1.1}cm \\ \hline
\texttt{Apartment\_2}    & $1.5$cm   & $1.0$cm & $0.9$cm & \best{0.9}cm \\ \hline
\texttt{Frl\_apartment\_2}    & $3.0$cm   & $2.8$cm & $2.8$cm & \best{1.9}cm \\ \hline
\texttt{Office\_0}    & $1.1$cm   & $0.9$cm & $0.7$cm & \best{0.7}cm \\ \hline
\texttt{Office\_1}    & $1.7$cm   & $1.2$cm & $1.8$cm & \best{1.4}cm \\ \hline
\texttt{Office\_2}    & $5.6$cm   & $140.4$cm & $8.3$cm & \best{4.3}cm \\ \hline
\texttt{Office\_3}    & $14.7$cm   & $13.5$cm & $9.2$cm & \best{8.0}cm \\ \hline
\texttt{Hotel\_0}    & $2.0$cm   & $160.4$cm & $1.8$cm & \best{1.5}cm \\ \hline
\texttt{Room\_0}    & $2.8$cm   & $3.5$cm & $3.3$cm & \best{2.3}cm \\ \hline
\texttt{Room\_1}    & $7.8$cm   & $4.4$cm & $4.5$cm & \best{4.0}cm \\ \hline
\end{tabular}
}}
\label{tab:ablt-tsdf}
\end{table} 

\paragraph{TSDF normalization -- \Sec{impl} -- \Eq{normallh}.}
We evaluate the necessity of TSDF normalization in likelihood estimation, which is composed of three design choices:
1) the determination of overlapping pixel set $O^t$ (\Eq{validness}), 2) the normalization of the summation of TSDF by set $O^t$, and 3) the adoption of the Kahan algorithm~\cite{neumaier74} in computing the summation of TSDF.
We test our method without each one of the three components:
\begin{compactitem}
\item \textbf{No Overlap}: The normalization of TSDF summation is performed over the full frame of depth map, i.e., replacing $O^t$ by $I^t_d$ in \Eq{normallh}.
\item \textbf{No Normalize}: The likelihood is estimated without normalizing the summation of TSDF, i.e., replacing \Eq{normallh} by \Eq{ourlh}.
\item \textbf{No Kahan}: The summation of TSDF is conducted directly, i.e., without using the Kahan algorithm.
\end{compactitem}
The evaluation is conducted on both ordinary sequences \DSICL and our fast-motion ones \DSFCMS. \Tab{ablt-tsdf} compares the tracking accuracy of our method and the various baselines.
It can be seen that ``No Overlap'' causes the most accuracy drop among the three baselines, suggesting its importance in likelihood estimation. For \texttt{Office\_2} and \texttt{Hotel\_0}, ``No Normalize'' produces large tracking error which can be attributed to exactly the cases of \Fig{overlap}(d), i.e., a very small portion of depth map lying within the valid range of TSDF due to very large rotation.
Although relatively less significant, ``No Kahan'' does affect the final tracking accuracy, hinting that numerical precision is a non-ignorable factor.

\begin{figure}[t]
\centering
\begin{overpic}
[width=0.482\linewidth]{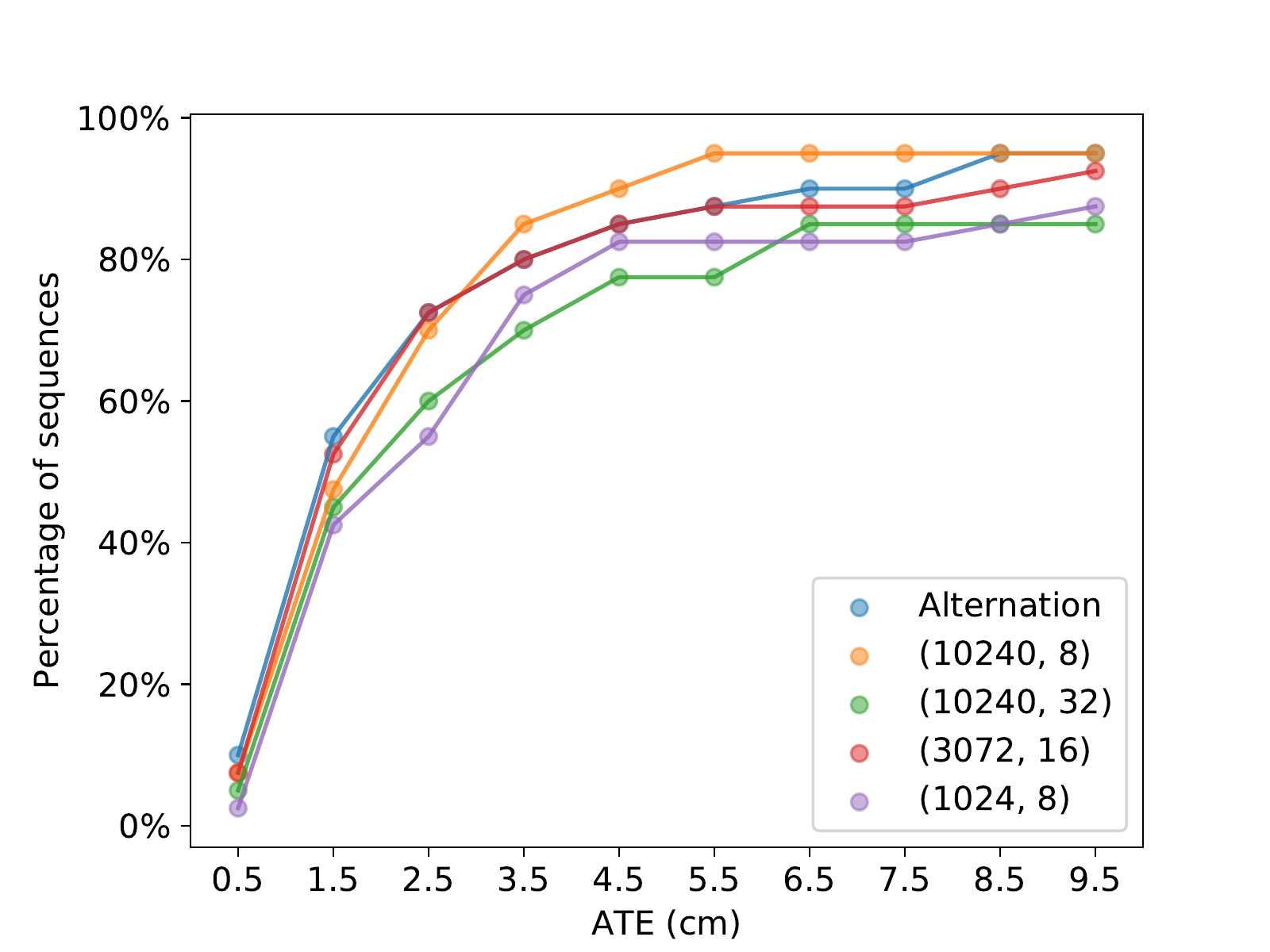}
\end{overpic}
\begin{overpic}
[width=0.496\linewidth]{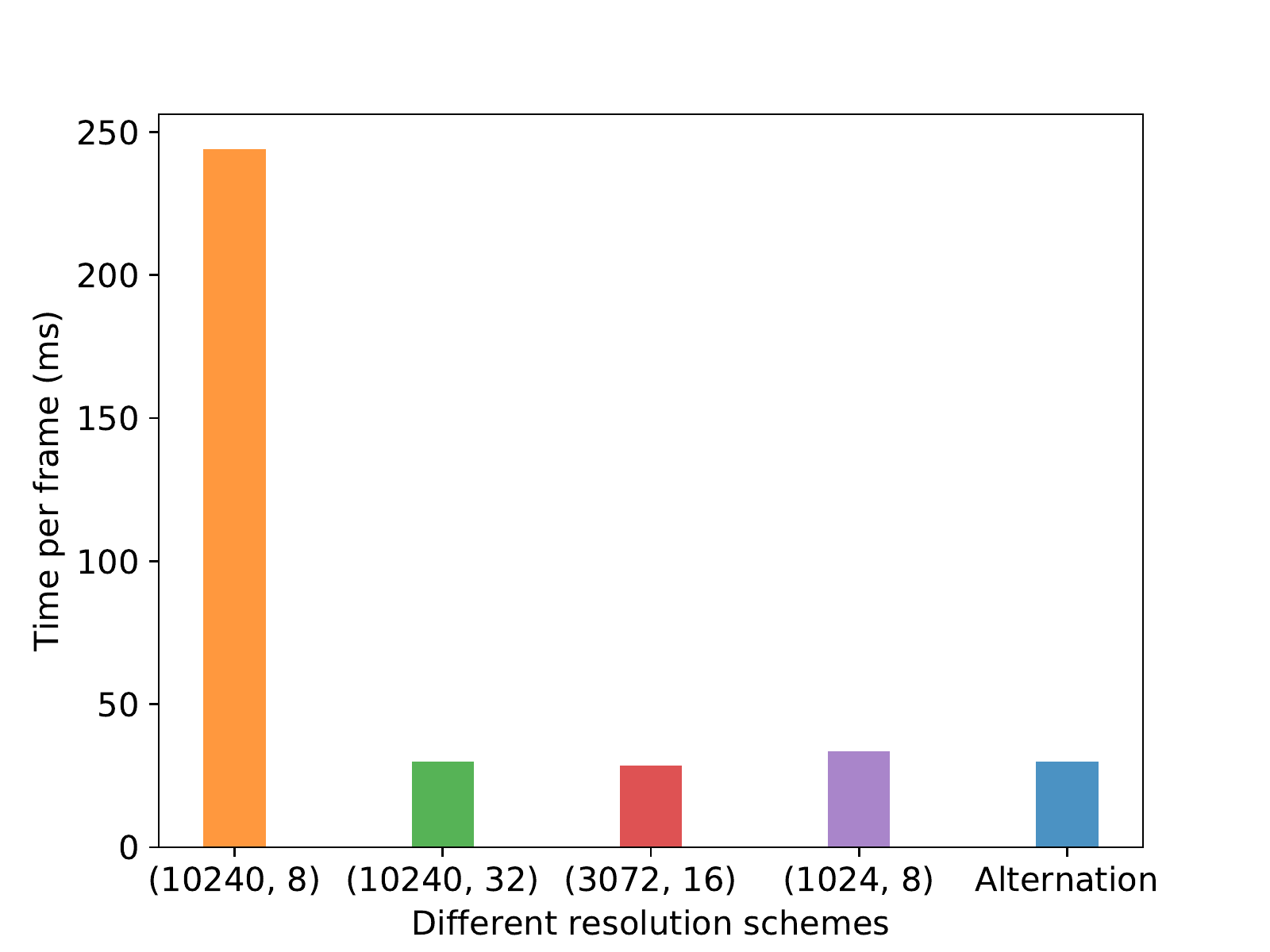}
\end{overpic}
\caption{
Comparing tracking accuracy and efficiency of our method (alternation between three resolution combinations) against various single resolution settings. Our method achieves the highest accuracy (success rate for different ATE thresholds) with a comparable cost to low-res the settings.
}
\label{fig:plot-multires}
\end{figure}

\paragraph{Multi-res PST alternation -- \Sec{impl}.}
To balance between the particle density of PST and the computational cost of fitness evaluation, we opt to work with three different combinations of PST resolution and depth map down-sampling rate: ($1024$, $1/8$), ($3072$, $1/16$), and ($10240$, $1/32$). The three combinations are used alternatively across iterations.
In \Fig{plot-multires}, we evaluate such multi-res alternation scheme by comparing the accuracy and efficiency of camera tracking against various single resolution options on \DSFCMS.
From the figure, the success rate of our alternation scheme is close to or higher than those high-res settings for almost all ATE thresholds, while our computational cost is at the same level as the low-res options. This verifies the effectiveness of our alternation scheme. It is the nature of swarm-guided PFO that enables this improvement: It is unnecessary to use a high-res PST at every step since 1) the best solution found by high-res PSTs would be passed along and inherited by the following iteration steps, and 2) the high-res depth maps associated with low-res PSTs yields more noise-robust likelihood estimation.

\begin{figure}[t]
\centering
\begin{overpic}
[width=0.325\linewidth]{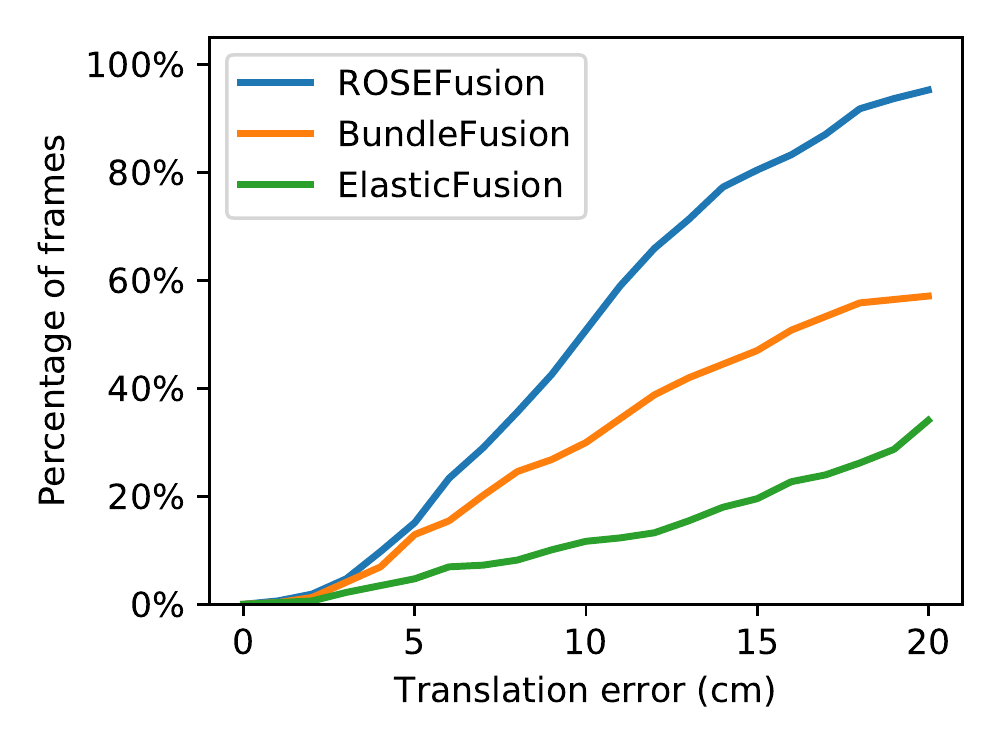}
\put(27,76){\footnotesize \texttt{camera\_shake\_1}}
\end{overpic}
\begin{overpic}
[width=0.325\linewidth]{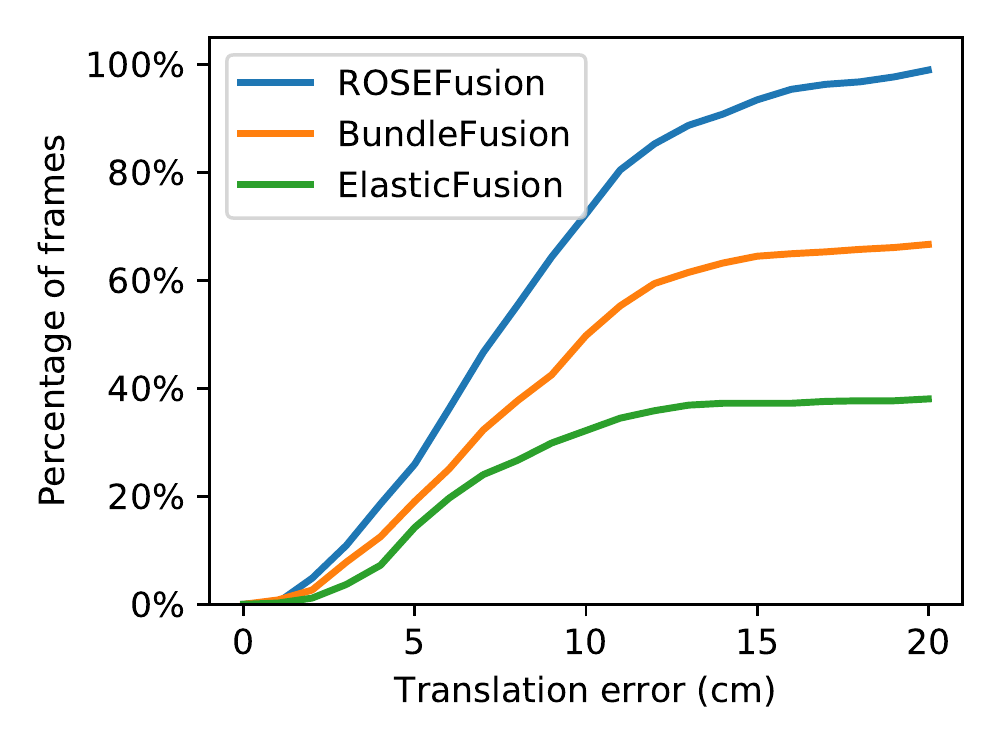}
\put(27,76){\footnotesize \texttt{camera\_shake\_2}}
\end{overpic}
\begin{overpic}
[width=0.325\linewidth]{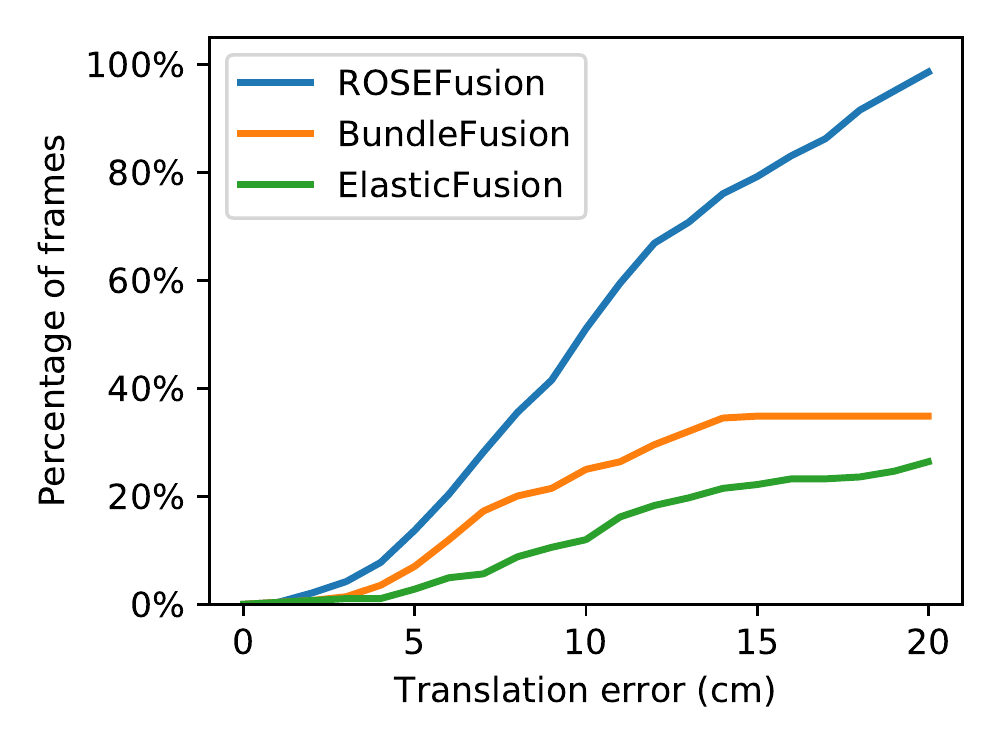}
\put(27,76){\footnotesize \texttt{camera\_shake\_3}}
\end{overpic}
\caption{
Comparing percentage of frames under increasing tolerance of translation error of per-frame pose on the three camera shaking sequences of \DSETH. Our method achieves much higher success rate of per-frame pose tracking than BundleFusion and ElasticFusion.
}
\label{fig:comp-plot-eth3d}
\end{figure}

\subsection{Quantitative Comparisons}
We quantitatively evaluate our method against several state-of-the-art methods for both ordinary and fast-motion sequences.

\paragraph{Comparison on ordinary benchmarks.}
In most existing datasets, the camera motion is usually slower than 1m/s. For those sequences, our method is able to achieve a comparable accuracy of camera tracking.
\Tab{comp-icl} compares ATE on the four sequences of \DSICL between our method and the state-of-the-art online (DVO-SLAM~\cite{kerl2013dense}, RGB-D SLAM~\cite{endres2012evaluation}, MRSMap~\cite{stuckler2014multi}, Kintinuous~\cite{Whelan2012}, VoxelHashing~\cite{Niessner2013}, ElasticFusion~\cite{Whelan2015}, BundleFusion~\cite{Dai2017}) and offline (Redwood~\cite{choi2015robust}) methods.
Our method achieves comparable accuracy to the best-performing method BundleFusion which involves a global pose optimization by bundle adjustment.
The \texttt{kt3} sequence is the most challenging one where there is barely no color or geometric features in some frames. Our method attains a decent accuracy even without any global optimization.
Similar comparative results can also be observed for \DSTUM; \emph{see the supplemental material}.

\begin{table}[!t]\centering
\caption{
Comparing the accuracy (ATE) of camera tracking on the four RGB-D sequences of \DSICL. The best and the second best results for each sequence are highlighted in \best{blue} and \sbest{green} colors, respectively.
}\vspace{-5pt}
\scalebox{0.95}{
\setlength{\tabcolsep}{1.6mm}{
\begin{tabular}{l|c|c|c|c}
\hline
Sequence           & \texttt{kt0} & \texttt{kt1} & \texttt{kt2} & \texttt{kt3}  \\ \hline\hline
DVO SLAM   & $10.4$cm   & $2.9$cm   & $19.1$cm  & $15.2$cm \\ \hline
RGB-D SLAM   & $2.6$cm   & $0.8$cm   & $1.8$cm  & $43.3$cm \\ \hline
MRSMap       & $20.4$cm   & $22.8$cm   & $18.9$cm  & $109$cm \\ \hline
Kintinuous    & $7.2$cm   & \sbest{0.5}cm   & \sbest{1.0}cm  & $35.5$cm \\ \hline
VoxelHashing  & $1.4$cm   & \best{0.4}cm   & $1.8$cm  & $12.0$cm \\ \hline
ElasticFusion   & $0.9$cm   & $0.9$cm   & $1.4$cm  & $10.6$cm \\ \hline
Redwood (rigid) & $25.6$cm   & $3.0$cm   & $3.3$cm  & $6.1$cm \\ \hline
BundleFusion    & \best{0.6}cm   & \best{0.4}cm   & \best{0.6}cm  & \best{1.1}cm \\ \hline
RoseFusion      & \sbest{0.8}cm   & $0.7$cm   & \sbest{1.0}cm  & \sbest{4.5}cm \\ \hline
\end{tabular}
}}
\label{tab:comp-icl}
\end{table} 

\begin{table}[!t]\centering
\caption{
Comparing the accuracy (ATE) of camera tracking on the three challenging RGB-D sequences of \DSETH. The best and the second best results for each sequence are highlighted in \best{blue} and \sbest{green} colors, respectively. `--' indicates that the tracking was failed.
}\vspace{-5pt}
\scalebox{0.95}{
\setlength{\tabcolsep}{0.4mm}{
\begin{tabular}{l|c|c|c}
\hline
Sequence           & \texttt{camera\_shake\_1} & \texttt{camera\_shake\_2} & \texttt{camera\_shake\_3}  \\ \hline\hline
BAD SLAM           & --   & --   & --   \\ \hline
DVO-SLAM           & $9.40$cm   & -- & -- \\ \hline
ORB-SLAM2          & --  & $6.89$cm  &  -- \\ \hline
ElasticFusion      & $8.44$cm   & -- & -- \\ \hline
BundleFusion       & \sbest{5.17}cm   & \sbest{3.49}cm   & -- \\ \hline
RoseFusion         & \best{0.62}cm   & \best{1.35}cm   & \best{4.67}cm \\ \hline
\end{tabular}
}}
\label{tab:comp-eth}
\end{table} 

\paragraph{Comparison on fast-motion benchmark -- \DSETH.}
The advantage of our method is best reflected on fast-motion sequences.
We first conduct a comparison on the three camera shake sequences of \DSETHCS. As reported in \Tab{speed-info}, the average angular velocity of camera motion of these sequences is nearly $10$ times faster than \DSICL and \DSTUM. Our method is compared to BAD-SLAM~\cite{schops2019bad}, DVO-SLAM, ORB-SLAM2~\cite{mur2017orb}, ElasticFusion and BundleFusion. \DSETHCS contains IMU data which was not used by any of the methods being compared. The results are reported in \Tab{comp-eth}. Our method was able to reconstruct all the three sequences with an acceptable tracking accuracy while none of the other methods could succeed on all.
\texttt{camera\_shake\_3} is the fastest sequence on which all other method failed while ours achieves $4.6$cm ATE.

In the plots of \Fig{comp-plot-eth3d}, we provide a breakdown study of per-frame pose accuracy for each camera shake sequence. In particular, we measure the frame-wise pose error using Translation Error (TE) and plot
the percentage of frames whose pose TE is below different thresholds.
BundleFusion drops frames which are lost tracking. We count the dropped frames as failed for all TE thresholds since the mechanism of frame dropping in BundleFusion is more involved than TE. Therefore, the comparison with BundleFusion may not be absolutely fair; we provide the results serving only as a reference.
It can be seen that our method achieves consistently more accurate per-frame pose estimation on all the three sequences.

\begin{figure*}[t]
\centering
\begin{overpic}
[width=0.196\linewidth]{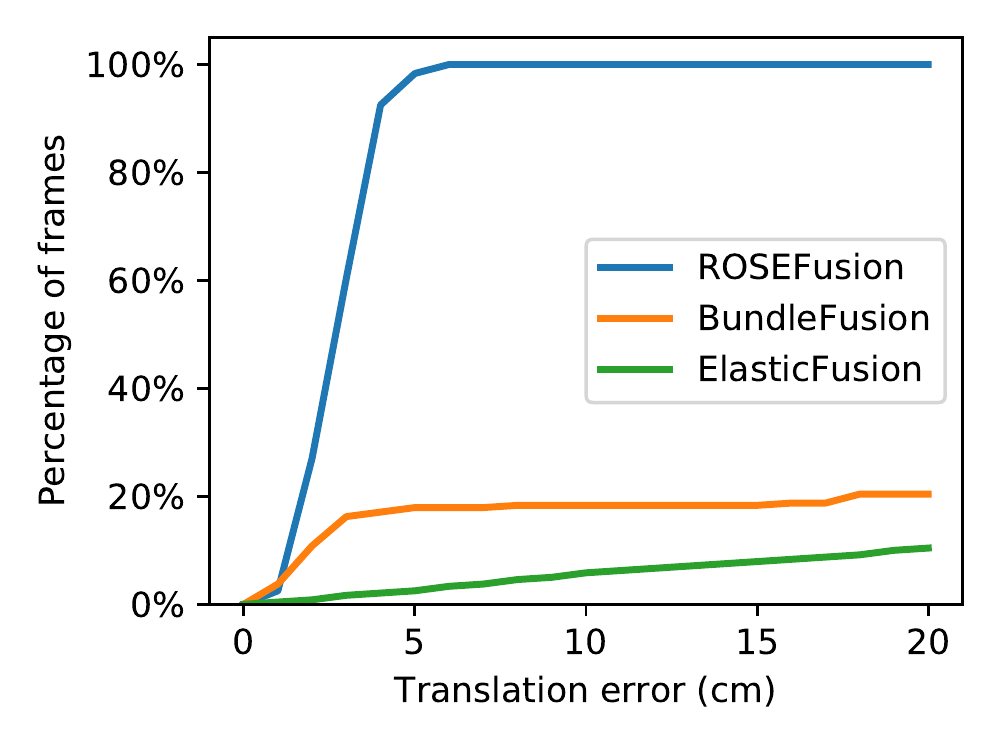}
\put(37,73){\footnotesize \texttt{apartment\_1}}
\end{overpic}
\begin{overpic}
[width=0.196\linewidth]{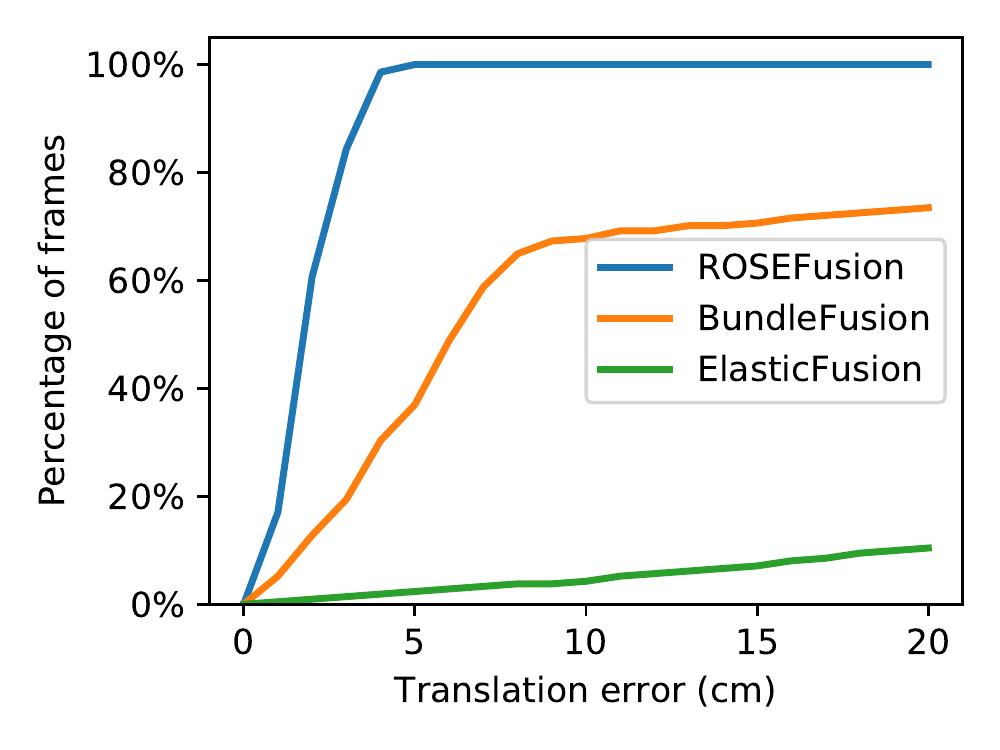}
\put(37,73){\footnotesize \texttt{apartment\_2}}
\end{overpic}
\begin{overpic}
[width=0.196\linewidth]{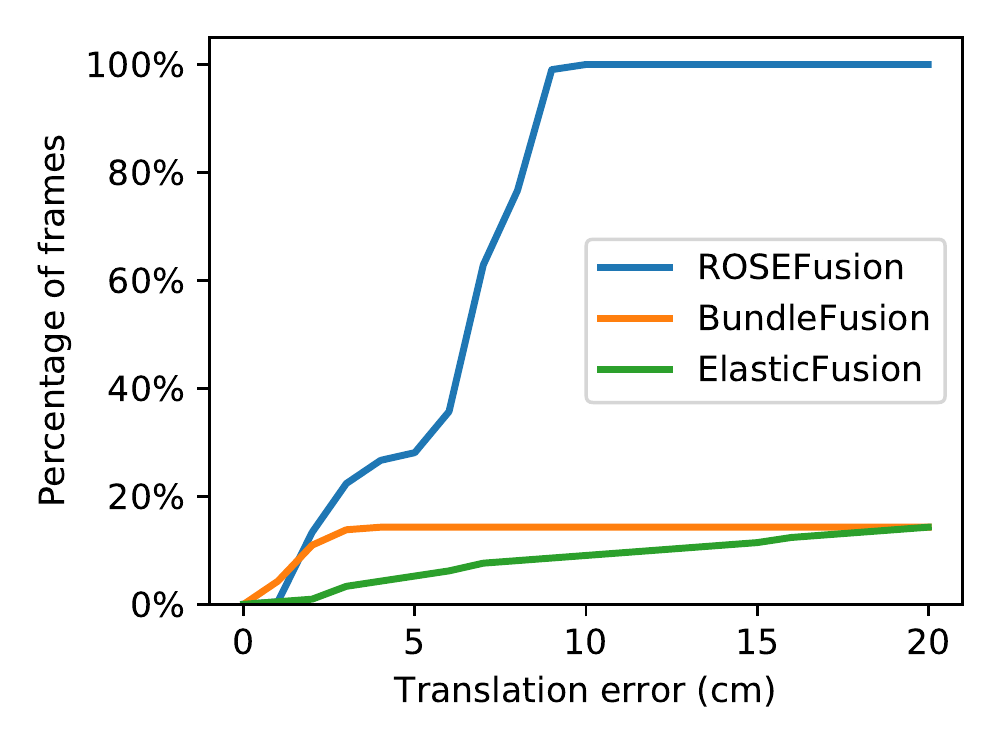}
\put(37,73){\footnotesize \texttt{apartment\_3}}
\end{overpic}
\begin{overpic}
[width=0.196\linewidth]{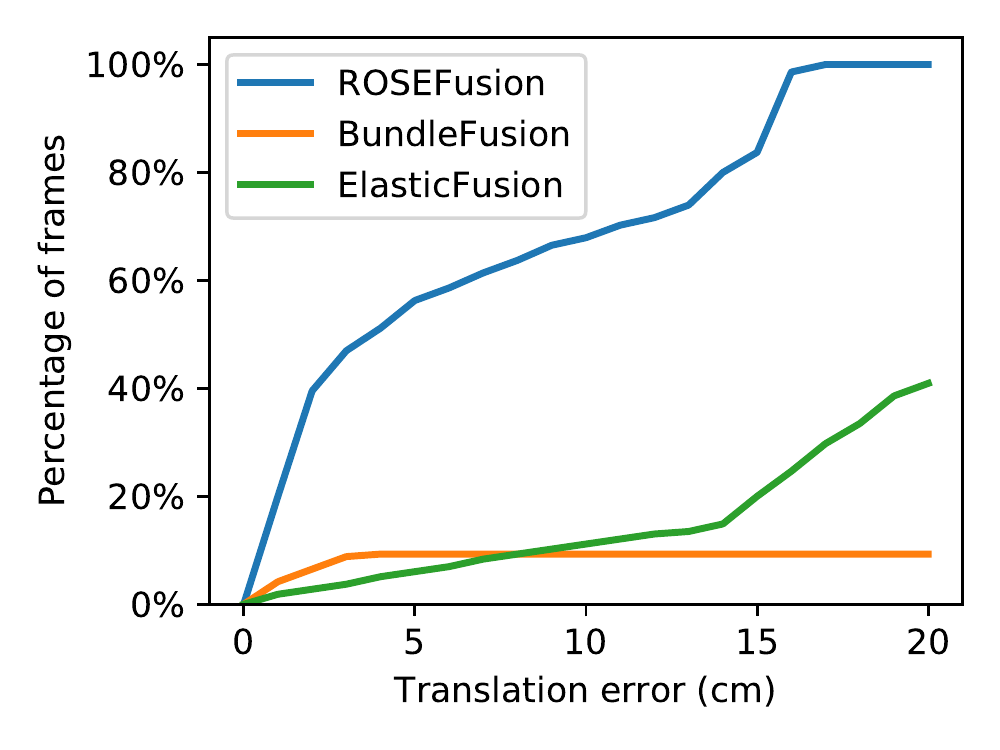}
\put(44,73){\footnotesize \texttt{hotel\_0}}
\end{overpic}
\begin{overpic}
[width=0.196\linewidth]{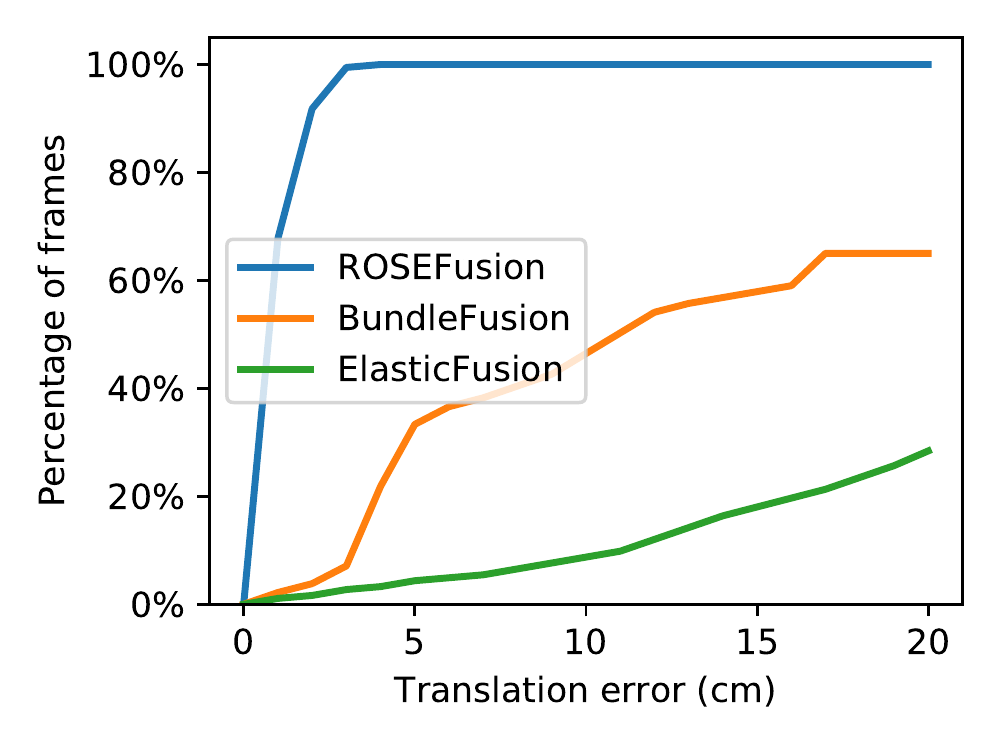}
\put(43,73){\footnotesize \texttt{office\_0}}
\end{overpic}
\begin{overpic}
[width=0.196\linewidth]{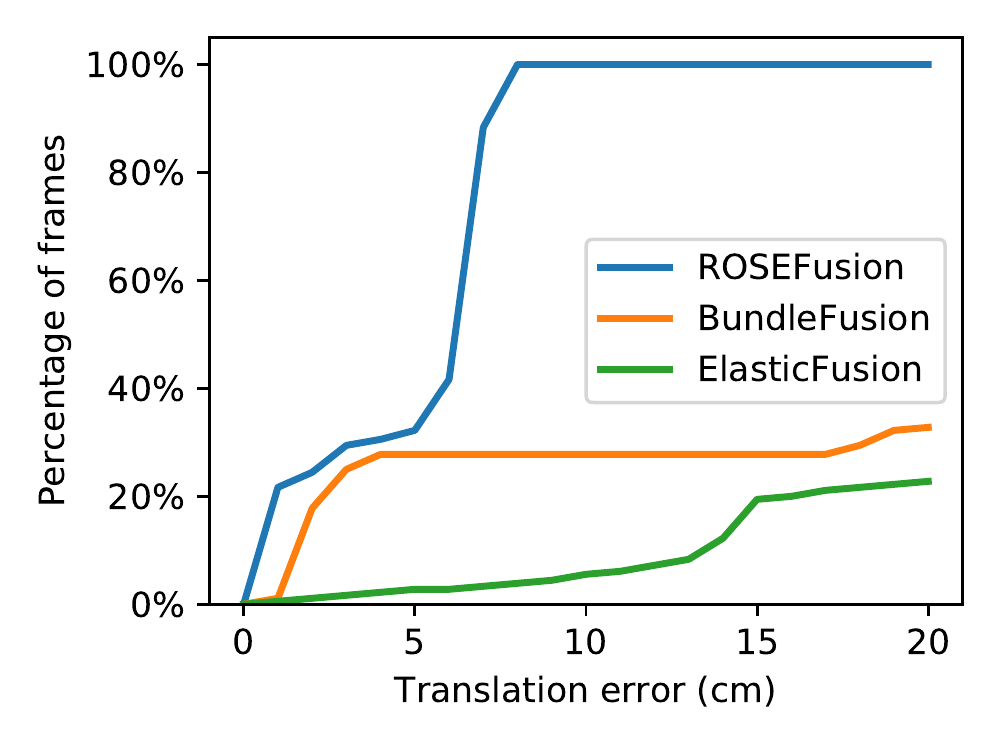}
\put(43,73){\footnotesize \texttt{office\_1}}
\end{overpic}
\begin{overpic}
[width=0.196\linewidth]{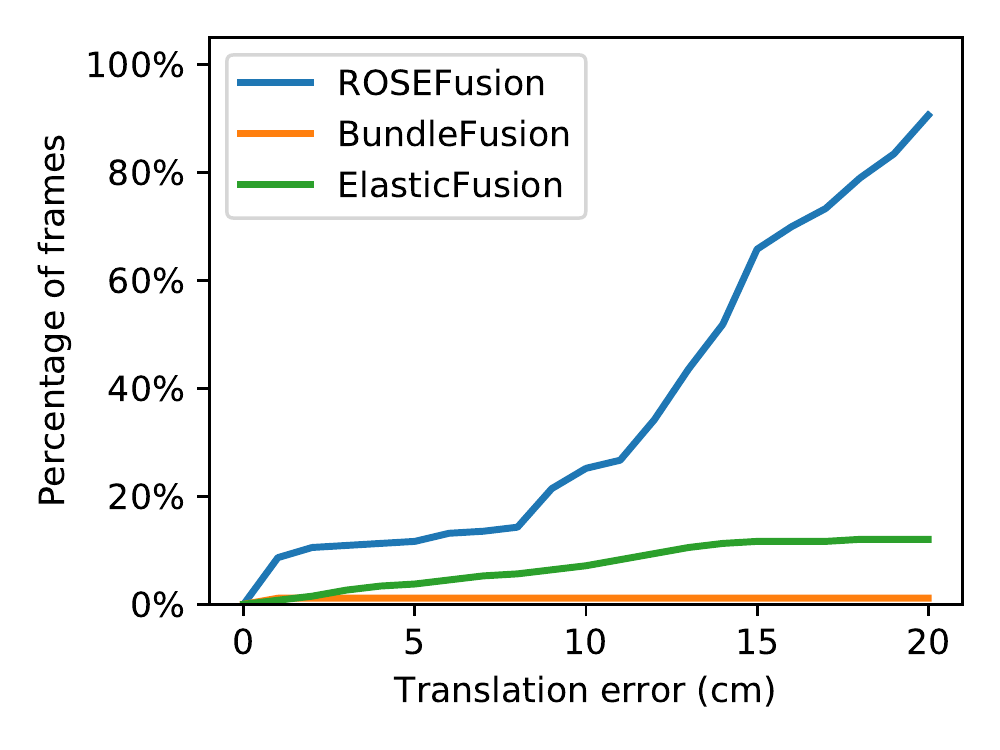}
\put(43,73){\footnotesize \texttt{office\_2}}
\end{overpic}
\begin{overpic}
[width=0.196\linewidth]{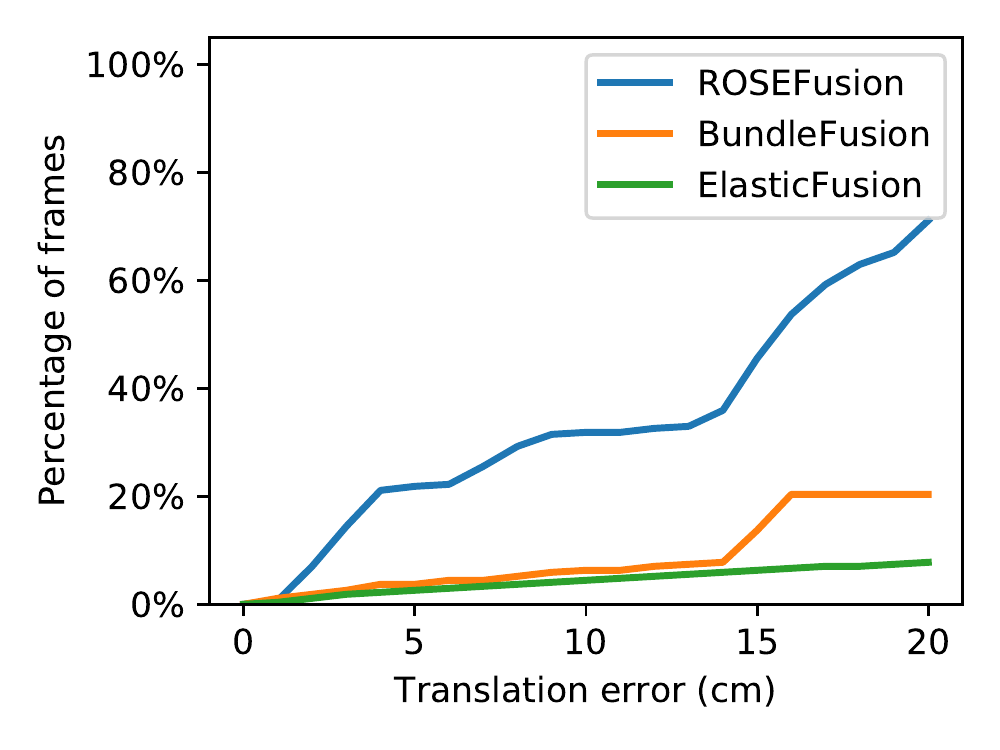}
\put(43,73){\footnotesize \texttt{office\_3}}
\end{overpic}
\begin{overpic}
[width=0.196\linewidth]{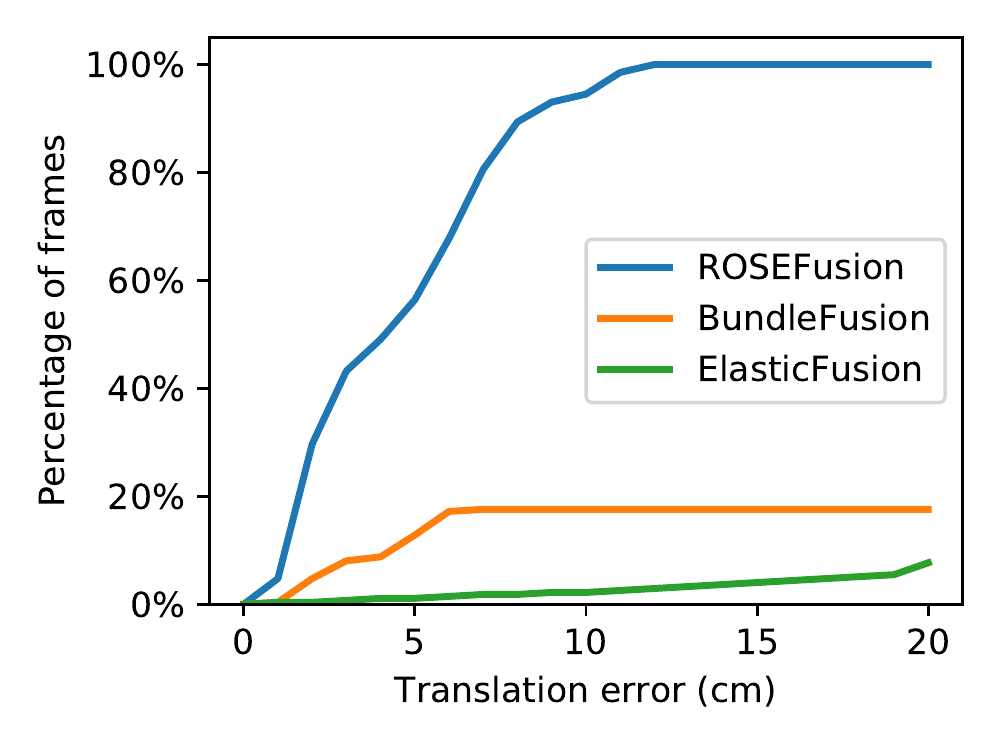}
\put(45,73){\footnotesize \texttt{room\_0}}
\end{overpic}
\begin{overpic}
[width=0.196\linewidth]{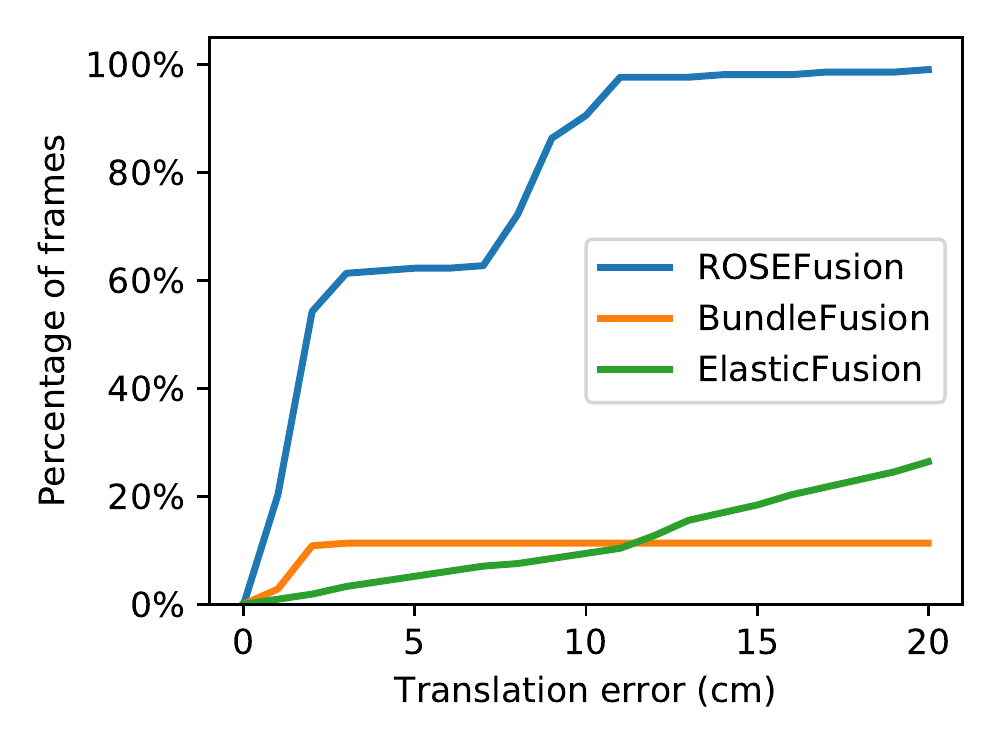}
\put(45,73){\footnotesize \texttt{room\_1}}
\end{overpic}
\caption{
Comparing percentage of frames under increasing tolerance of translation error of per-frame pose on ten sequences of our \DSFCMS dataset. Our method achieves significantly higher success rate of per-frame pose tracking than BundleFusion and ElasticFusion.
}
\label{fig:comp-plot-fcm}
\end{figure*}

\paragraph{Comparison on fast-motion benchmark -- \DSFCM.}
\Tab{comp-fcm-synth} provides a comparison on \DSFCMS. Based on the ground-truth trajectories of this synthetic dataset, we compare our method with ElasticFusion and BundleFusion on the ten sequences. Note that, in what follows, we mainly compare to these two methods in the various fast-motion tests since they are the representative state-of-the-arts in point-based and volumetric fusion, respectively.

Compare to the two baselines, our method demonstrates significantly higher tracking accuracy for all the sequences except \texttt{Room\_1}. The \texttt{Room\_1} sequence is quite challenging since the empty room contains barely no color or geometric feature. BundleFusion is able to reconstruct a small portion of the room with good tracking accuracy while dropping $88.8\%$ of the frames. On the contrary, our method and ElasticFusion do not drop frame. Similar situation is also found in \texttt{Room\_0}, \texttt{Office\_2} and \texttt{Office\_3}, on which our method achieves successful reconstruction. \Fig{gallery-synth} shows three examples of visual reconstruction results for \DSFCMS.

\begin{table}[!t]\centering
\caption{
Comparing the accuracy (ATE) of camera tracking on the ten scenes of \DSFCMS. The best results for each sequence are highlighted in \best{blue} color. `--' indicates that the tracking was failed. The frame dropping rate of BundleFusion is high for these sequences (e.g., $88.8\%$ of the frames of \texttt{Room\_1} were dropped by BundleFusion) while ElasticFusion and our method did not drop any frame. \kx{Note, however, ATE is reported for reconstructed frames only for all methods being compared.}
}\vspace{-5pt}
\scalebox{0.95}{
\setlength{\tabcolsep}{1.0mm}{
\begin{tabular}{l|c|c|c}
\hline
Method               & ElasticFusion & BundleFusion & Ours \\ \hline\hline
\texttt{Apartment\_1}    & $40.9$cm   & $4.6$cm  & \best{1.1}cm \\ \hline
\texttt{Apartment\_2}    & $40.7$cm   & $2.2$cm  & \best{1.0}cm \\ \hline
\texttt{Frl\_apartment\_2}    & $43.8$cm   & $83.6$cm  & \best{1.9}cm \\ \hline
\texttt{Office\_0}    & $22.3$cm   & $2.7$cm  & \best{0.7}cm \\ \hline
\texttt{Office\_1}    & $2.3$cm   & $17.3$cm & \best{1.4}cm \\ \hline
\texttt{Office\_2}    &  $65.9$cm  & $93.0$cm  & \best{4.3}cm \\ \hline
\texttt{Office\_3}    & $94.3$cm   & $253.5$cm & \best{8.0}cm \\ \hline
\texttt{Hotel\_0}    & $43.8$cm   & $65.2$cm  & \best{1.5}cm \\ \hline
\texttt{Room\_0}    & --   & --  & \best{2.3}cm \\ \hline
\texttt{Room\_1}    & $31.0$cm   & \best{0.6}cm  & $4.0$cm \\ \hline
\end{tabular}
}}\vspace{-12pt}
\label{tab:comp-fcm-synth}
\end{table} 

\begin{table}[!t]\centering
\caption{Comparing reconstruction completeness (Compl.) and accuracy (Acc.) of ElasticFusion, BundleFusion and our method over the \DSFCMR dataset. The best results for each sequence are highlighted in \best{blue} color.
\kx{We used the default parameters of the two alternatives; better results could be obtained by tuning their parameters.}}
\vspace{-5pt}
\scalebox{0.95}{
\setlength{\tabcolsep}{1.2mm}{
\begin{tabular}{l|cccccc}
\hline
  & \multicolumn{2}{c|}{ElasticFusion} & \multicolumn{2}{c|}{BundleFusion} & \multicolumn{2}{c}{Ours}                     \\ \cline{2-7}
          & \multicolumn{1}{c|}{Compl.}  & \multicolumn{1}{c|}{Acc.}    & \multicolumn{1}{c|}{Compl.}  & \multicolumn{1}{c|}{Acc.}  & \multicolumn{1}{c|}{Compl.} & Acc.  \\ \hline\hline
\texttt{Apartment\_I}     & \multicolumn{1}{c|}{$22.1\%$}     & \multicolumn{1}{c|}{$7.7$cm}    & \multicolumn{1}{c|}{$34.2\%$}     & \multicolumn{1}{c|}{$6.4$cm}  & \multicolumn{1}{c|}{\best{84.3\%}}      & \best{4.8}cm    \\ \hline
\texttt{Apartment\_II}    & \multicolumn{1}{c|}{$15.0\%$}     & \multicolumn{1}{c|}{$7.2$cm}    & \multicolumn{1}{c|}{$25.2\%$}     & \multicolumn{1}{c|}{$5.2$cm}  & \multicolumn{1}{c|}{\best{86.6\%}}      & \best{4.2}cm    \\ \hline
\texttt{Lab}  & \multicolumn{1}{c|}{$15.1\%$}     & \multicolumn{1}{c|}{$7.3$cm}    & \multicolumn{1}{c|}{$16.9\%$}     & \multicolumn{1}{c|}{$5.4$cm}  & \multicolumn{1}{c|}{\best{91.6\%}}      & \best{4.8}cm    \\ \hline
\texttt{Stairwell} & \multicolumn{1}{c|}{$11.7\%$}     & \multicolumn{1}{c|}{$8.3$cm}    & \multicolumn{1}{c|}{$14.4\%$}     & \multicolumn{1}{c|}{$5.8$cm}  & \multicolumn{1}{c|}{\best{82.8\%}}      & \best{5.4}cm    \\ \hline
\texttt{Gym}     & \multicolumn{1}{c|}{$61.1\%$}     & \multicolumn{1}{c|}{$7.7$cm}    & \multicolumn{1}{c|}{$12.4\%$}     & \multicolumn{1}{c|}{\best{5.1}cm}  & \multicolumn{1}{c|}{\best{61.8\%}}      & $6.9$cm   \\ \hline
\texttt{Lounge\_I}      & \multicolumn{1}{c|}{$10.7\%$}     & \multicolumn{1}{c|}{$7.8$cm}    & \multicolumn{1}{c|}{$7.7\%$}     & \multicolumn{1}{c|}{$6.1$cm}  & \multicolumn{1}{c|}{\best{93.5\%}}      & \best{4.5}cm    \\ \hline
\texttt{Lounge\_II}      & \multicolumn{1}{c|}{$8.8\%$}     & \multicolumn{1}{c|}{$9.0$cm}    & \multicolumn{1}{c|}{$1.8\%$}     & \multicolumn{1}{c|}{$9.4$cm}  & \multicolumn{1}{c|}{\best{87.9\%}}      & \best{5.4}cm    \\ \hline
\texttt{Studio}      & \multicolumn{1}{c|}{$36.1\%$}     & \multicolumn{1}{c|}{$7.2$cm}    & \multicolumn{1}{c|}{$63.8\%$}     & \multicolumn{1}{c|}{\best{4.9}cm}  & \multicolumn{1}{c|}{\best{65.1\%}}      & \best{4.9}cm    \\ \hline
\texttt{Meeting\_room}      & \multicolumn{1}{c|}{$17.9\%$}     & \multicolumn{1}{c|}{$8.5$cm}    & \multicolumn{1}{c|}{$20.1\%$}     & \multicolumn{1}{c|}{$6.9$cm}  & \multicolumn{1}{c|}{\best{90.0\%}}      & \best{5.8}cm    \\ \hline
\texttt{Office}      & \multicolumn{1}{c|}{$16.4\%$}     & \multicolumn{1}{c|}{$8.3$cm}    & \multicolumn{1}{c|}{$6.1\%$}     & \multicolumn{1}{c|}{$5.6$cm}  & \multicolumn{1}{c|}{\best{67.7\%}}      & \best{5.4}cm    \\ \hline
\texttt{Workshop\_I}      & \multicolumn{1}{c|}{$11.6\%$}     & \multicolumn{1}{c|}{$9.2$cm}    & \multicolumn{1}{c|}{$25.4\%$}     & \multicolumn{1}{c|}{\best{5.6}cm}  & \multicolumn{1}{c|}{\best{55.9\%}}      & $5.7$cm    \\ \hline
\texttt{Workshop\_II}      & \multicolumn{1}{c|}{$16.3\%$}     & \multicolumn{1}{c|}{$7.0$cm}    & \multicolumn{1}{c|}{$51.6\%$}     & \multicolumn{1}{c|}{\best{5.4}cm}  & \multicolumn{1}{c|}{\best{66.4\%}}      & \best{5.4}cm    \\ \hline
\end{tabular}
}}
\label{tab:comp-fcm-real}
\end{table} 

Similar to \Fig{comp-plot-eth3d}, \Fig{comp-plot-fcm} provides breakdown analyses of frame-wise pose accuracy for the ten sequences. These plots also reflect the difficulty of the above-mentioned sequences; see the slowly growing percentage. Our method again performs consistently better than the two baseline methods. For \texttt{Room\_1}, our curve is notably higher than BundleFusion indicating that our method attains an overall better tracking when all frames are counted in.

In \Tab{comp-fcm-real}, we also conduct a comparison on $12$ real captured sequences of \DSFCMR. \supl{See the results on the other $12$ sequences in the supplemental material.} Since these sequences only possess ground-truth reconstruction by LiDAR, we evaluate the reconstruction quality, i.e., the completeness and accuracy w.r.t. the ground-truth surfaces. As the reconstruction accuracy measures the RMS error over only the overlapping (inlier) regions between the reconstructed and the ground-truth surfaces, the numbers of the three methods are close to each other \kx{although ours does not involve any post-processing of global pose optimization or loop closure as the two alternative. In \Tab{comp-fcm-real}, we set the threshold of inlier to $15$cm. When the threshold is set to $5$cm, the average error is $1\sim3$cm, with about $10\%$ drop in completeness.} The reconstruction quality is best exposed by completeness on which our method is consistently and significantly better than the two alternatives. The visual results of reconstruction for this dataset can be found in \Fig{gallery}.

\begin{figure}[t]
\centering
\begin{overpic}
[width=0.90\linewidth]{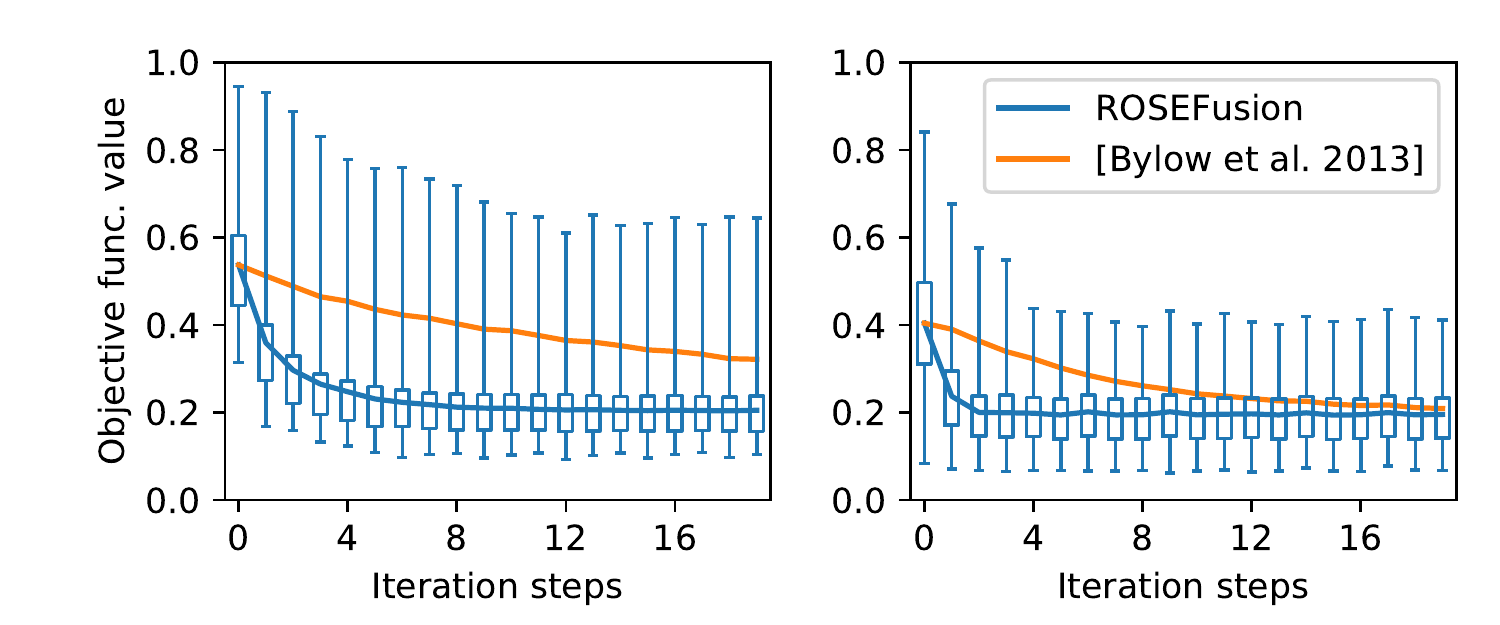}
\put(18,41){\footnotesize \DSFCMS}
\put(72,41){\footnotesize \DSICL}
\end{overpic}
\caption{
Plots of average objective function values at different iteration steps for our method (blue) and~\cite{bylow2013real} (orange). For our method, we also show the range of objective of all particles over all frames. The results are reported for both \DSFCMS (left) and \DSICL (right).
}
\label{fig:comp-plot-rss}
\end{figure}

\paragraph{Comparison with \cite{bylow2013real}.}
Our method is closely related to~\cite{bylow2013real} in which a similar objective was defined based on the summation of TSDF over the unprojected 3D points of a depth map (\Eq{nll}). This makes their method correspondence-free too. Different from our method, however, they pursue a gradient descent approach and employ the Gauss-Newton method to minimize the summation of TSDF. This nonlinear least square approach finds difficulty in handling large rotation as shown in many related works~\cite{huang2006subspace}. Moreover, the gradient can be undefined when the unprojected 3D points lie out of the valid range of TSDF when the camera undergoes a large transformation.

In \Fig{comp-plot-rss} (left column), we plot the objective function values (\Eq{nll}; the lower the better) of all frames and all sequences at different iteration steps.
For our method, we plot both the average values and the range of values of all particles.
For their method, we simply plot the actual objective function values at each iteration step.
Their values are mostly larger than the medium half of our particles (see the slim boxes).
This demonstrates that our method minimizes the objective with a significantly faster convergence for both of the two datasets.

\subsection{Qualitative Results}
\label{sec:qualitative}

\begin{figure}[t]
\centering
\begin{overpic}
[width=\linewidth]
{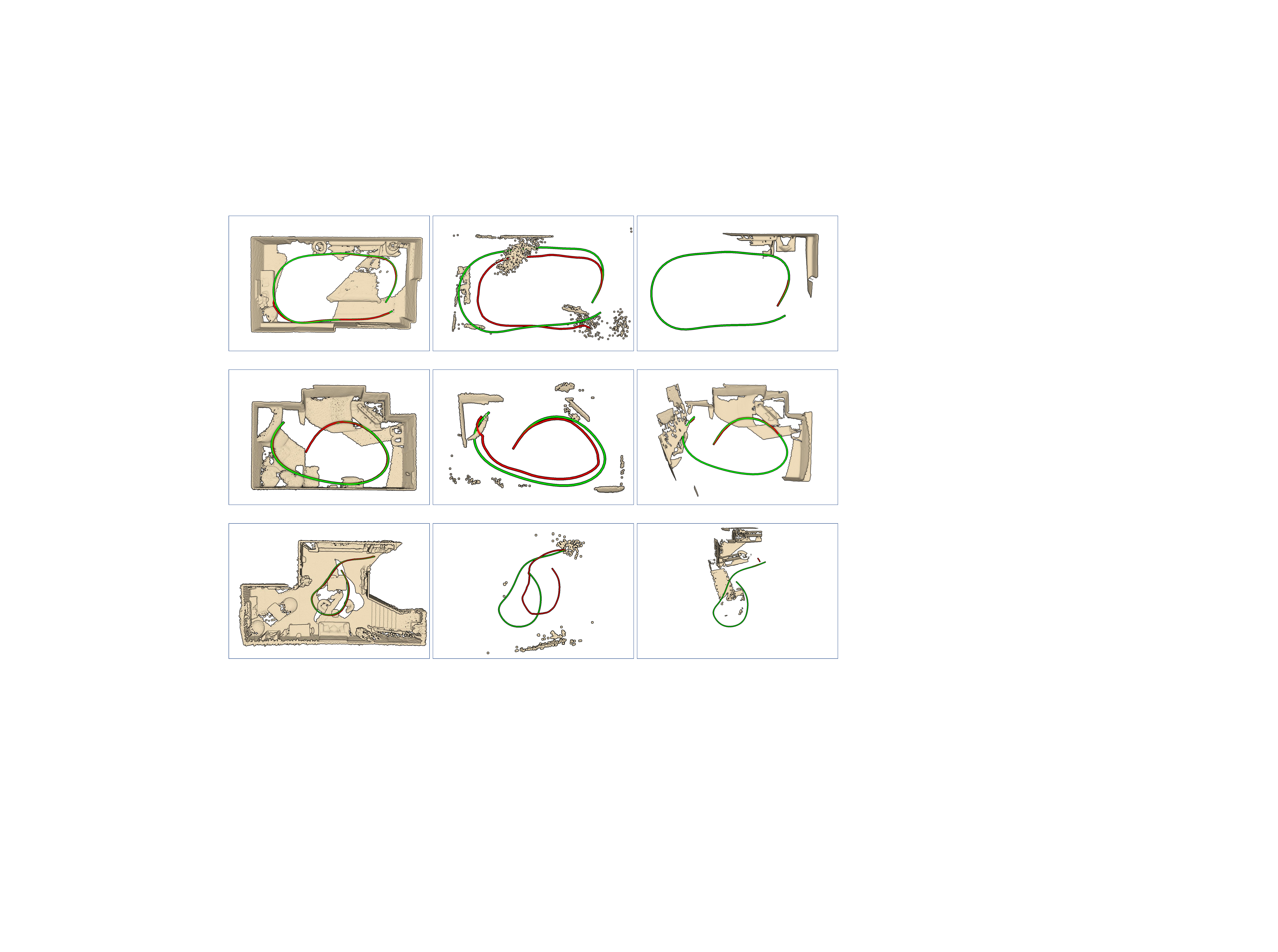}
\put(9,74){\footnotesize ROSEFusion}
\put(43,74){\footnotesize ElasticFusion}
\put(75,74){\footnotesize BundleFusion}
\put(.5,70.5){\footnotesize \texttt{Room\_1}}
\put(.5,45.5){\footnotesize \texttt{Office\_1}}
\put(.5,20.2){\footnotesize \texttt{Frl\_apartment\_2}}
\end{overpic}
\caption{
3D reconstruction results and tracked camera trajectories (red curves) for three fast-motion sequences of \DSFCMS. For each sequence, we compare the results of ROSEFusion (left), ElasticFusion (middle) and BundleFusion (right). The ground-truth trajectories (green curves) are overlaid for reference purpose.
}
\label{fig:gallery-synth}\vspace{-8pt}
\end{figure}
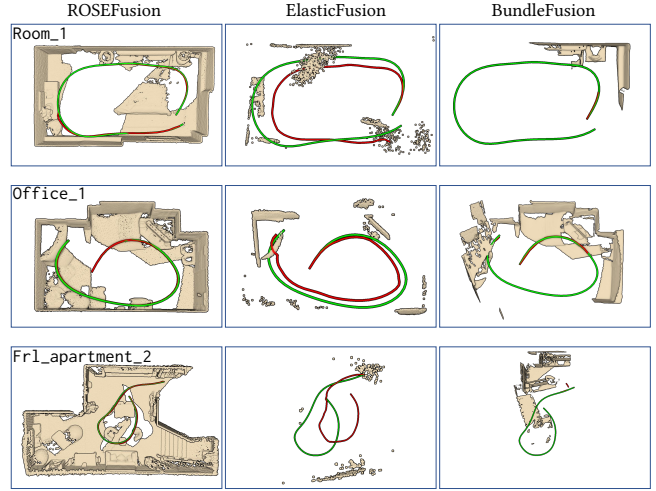

\paragraph{Visual comparison of reconstruction.}
We provide visual results on both synthetic and real fast-motion datasets.
\Fig{gallery-synth} shows the reconstructions on three sequences of \DSFCMS (\supl{see the full set of results in the supplemental material}). For each sequence, we show the reconstruction results along with the tracked camera trajectories for ROSEFusion (left), ElasticFusion (middle) and BundleFusion (right).
The tracked trajectories are overlaid on top of the ground-truth ones for visual contrasting. It can be seen that our trajectory matches well to the ground-truth. For BundleFusion, the trajectory segments corresponding to those remaining frames (did not get dropped) also conform well against the ground-truth.

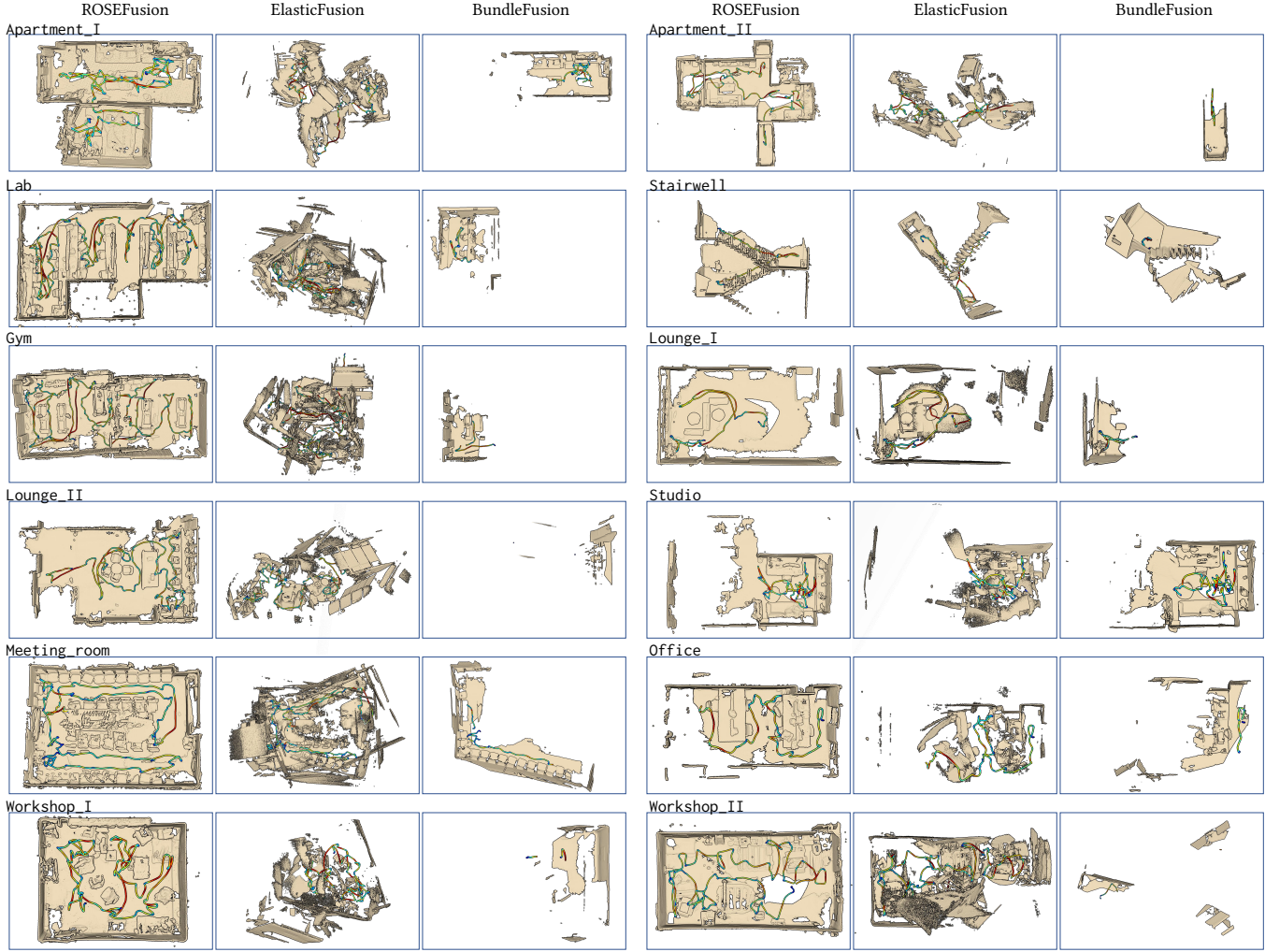
\begin{figure*}[t]
\centering
\begin{overpic}
[width=\linewidth]
{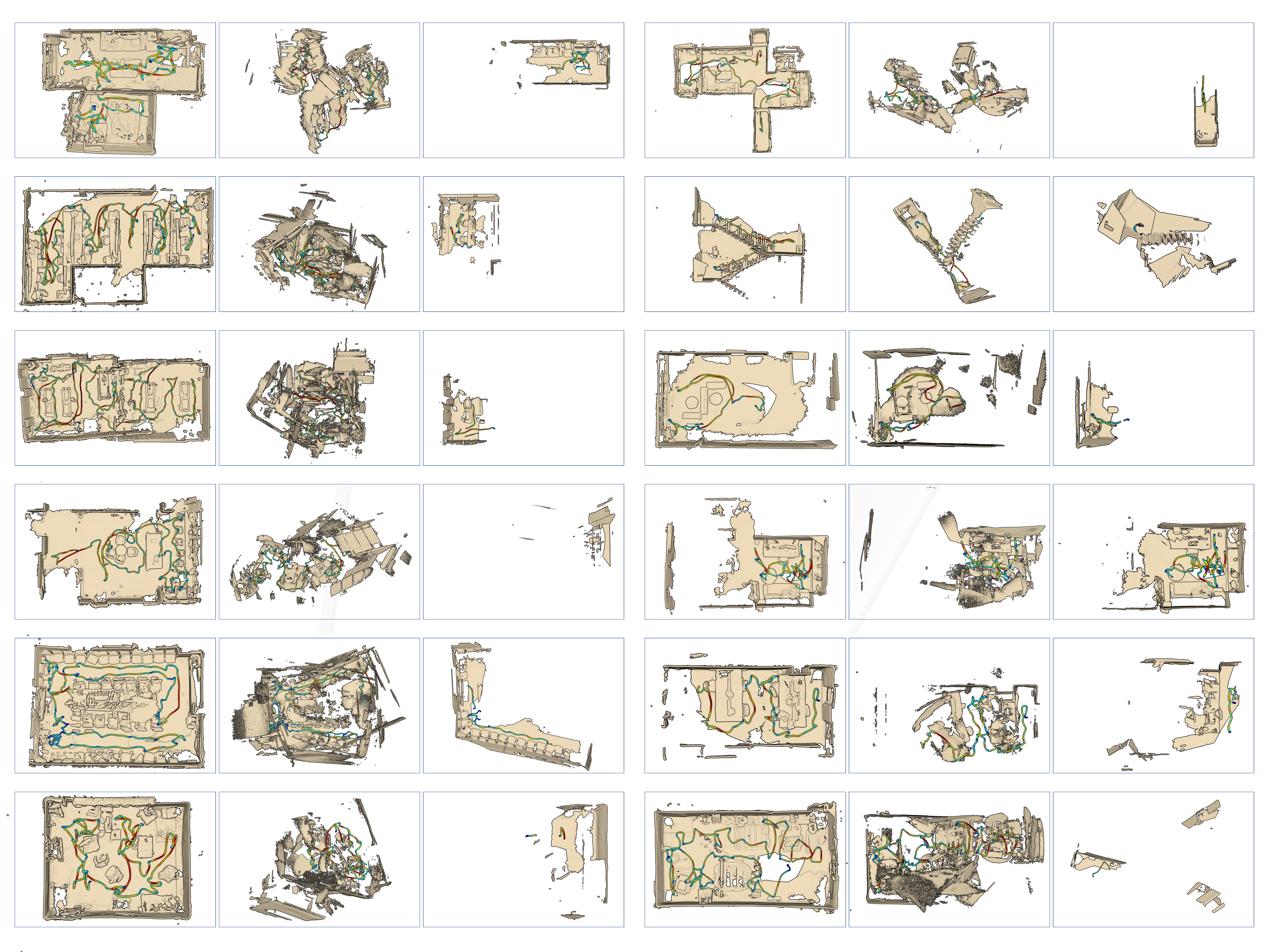}
\put(6,74.5){\footnotesize ROSEFusion}
\put(21,74.5){\footnotesize ElasticFusion}
\put(37,74.5){\footnotesize BundleFusion}
\put(56,74.5){\footnotesize ROSEFusion}
\put(72,74.5){\footnotesize ElasticFusion}
\put(88,74.5){\footnotesize BundleFusion}
\put(0,73){\footnotesize \texttt{Apartment\_I}}
\put(51,73){\footnotesize \texttt{Apartment\_II}}
\put(0,60.6){\footnotesize \texttt{Lab}}
\put(51,60.6){\footnotesize \texttt{Stairwell}}
\put(0,48.5){\footnotesize \texttt{Gym}}
\put(51,48.5){\footnotesize \texttt{Lounge\_I}}
\put(0,36){\footnotesize \texttt{Lounge\_II}}
\put(51,36){\footnotesize \texttt{Studio}}
\put(0,23.7){\footnotesize \texttt{Meeting\_room}}
\put(51,23.7){\footnotesize \texttt{Office}}
\put(0,11.3){\footnotesize \texttt{Workshop\_I}}
\put(51,11.3){\footnotesize \texttt{Workshop\_II}}
\end{overpic}
\caption{
Gallery of 3D reconstruction results along with tracked camera trajectories for the twelve real captured fast-motion sequences of \DSFCMR. For each sequence, we compare the results of ROSEFusion (left), ElasticFusion (middle) and BundleFusion (right). The linear velocity of camera movement is color-coded (refer to \Fig{teaser} for color bar) along the trajectories.
}
\label{fig:gallery}\vspace{-10pt}
\end{figure*}

\Fig{gallery} is a gallery of visual results for \DSFCMR, in a similar layout to \Fig{gallery-synth}.
The linear velocity of camera movement (estimated based on our tracking) is color-coded (refer to \Fig{teaser} for color bar) along the trajectories.
Note how our method is able to successfully reconstruct most of the sequences on which the other two methods failed. The \texttt{Stairwell} sequence is especially challenging since it contains repetitive structure but no prominent color features. Nevertheless, our method succeeds on it with good reconstruction quality. See also \Tab{comp-fcm-real} for quantitative measurement. The \texttt{Gym} sequence is also quite challenging due to the large mirror on the wall. The mirror reflectance causes drastically missing/erroneous depth such that our method produces some drift along the wall which is any loop closure or global optimization mechanism.

\begin{figure*}[t]
\centering
\begin{overpic}
[width=\linewidth]
{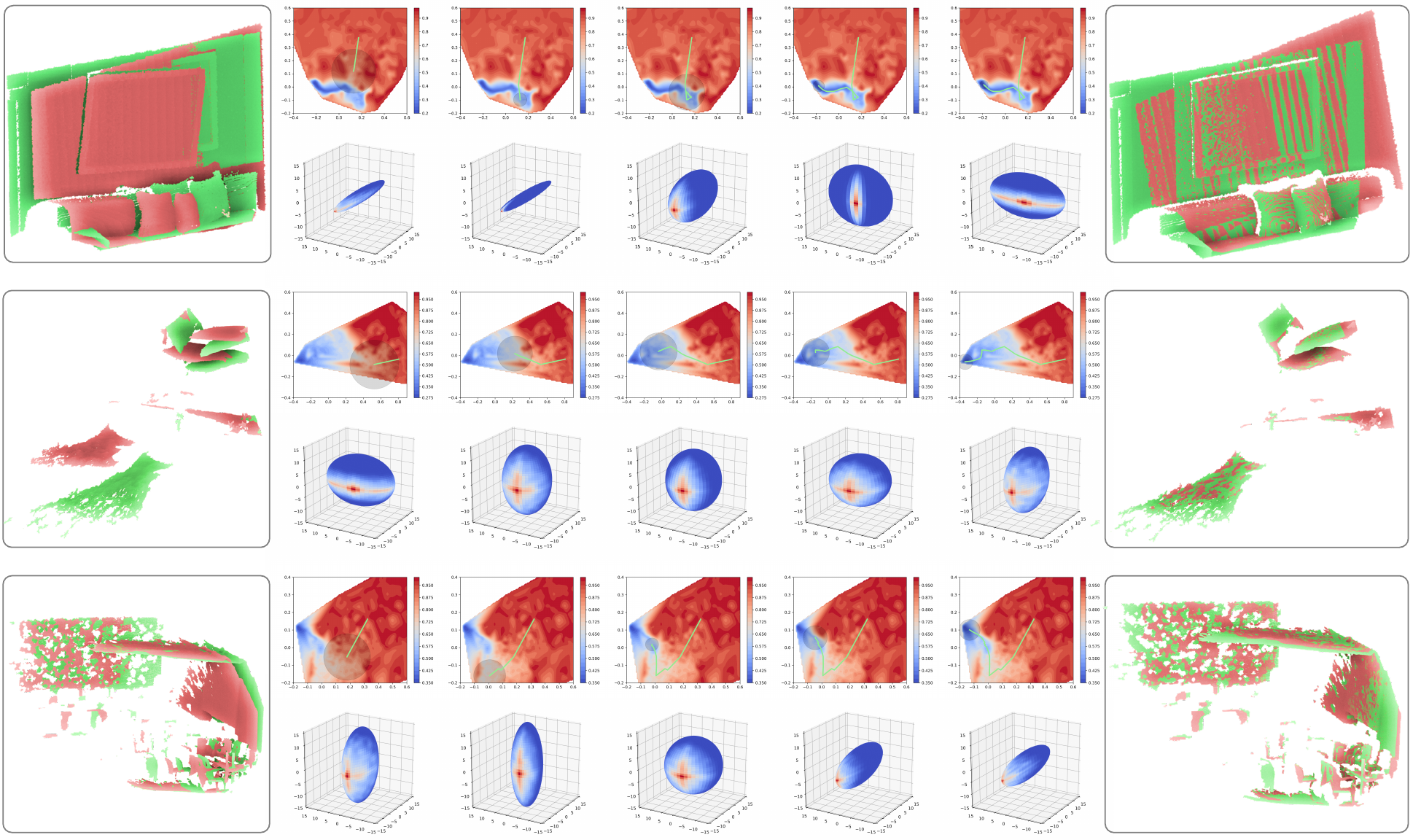}
\end{overpic}
\caption{
Visualization of pose optimization process on three representative frames selected from \texttt{kt3} of \DSICL, \texttt{camera\_shake\_3} of \DSETHCS and \texttt{Lab} of \DSFCMR, respectively. In each row, we show two frames with initial poses, the evolution of PST during optimization, and the final alignment result. The upper part of the evolution sequence shows the optimization landscape (colored plot) and the exploration path (green curve) of PST. The lower part shows the evolution of the 6D PST visualized with a 3D ellipsoid.
}
\label{fig:plot-3d-opt}
\end{figure*}

\paragraph{Visualization of optimization process.}
In \Fig{plot-3d-opt}, we present a visualization of the per-frame pose optimization process.
Given the current frame (the green point cloud) with an initial pose, the goal is to optimize its pose to align it against the previous frame (the red point cloud). For each example, the initial and final configurations are shown at the two ends and the progressive evolution of PST is visualized in-between. In the upper part of the evolution sequence, we visualize the landscape of objective (\Eq{nll}) via 2D Isomap embedding~\cite{tenenbaum2000global} of the 6D solution space, as well as the exploration path of our PST. The lower part shows the evolution of the 6D PST visualized as a 3D ellipsoid. The axis lengths of the 3D ellipsoid depict the rescaling factors along the three dimensions corresponding to translation and the color map encodes the scales along three rotational dimensions. The color map on the ellipsoid can be seen as the Gauss map of camera orientations of all particles. Note that the rotation here refers to relative rotation w.r.t. the pose of the previous step. Using relative rotation makes the color maps well aligned and the color change easy to observe.

The visualization clearly demonstrates the power of our random optimization scheme. Although the optimization landscape is highly non-convex due to large inter-frame transformations, our PFO guided by PST can always finds a good local optimum efficiently. It enables good quality frame-to-frame alignment without needing photometric or geometric feature detection and correspondence.

\subsection{Complexity and Runtime Analysis}
The time complexity of our random optimization is $O(NM)$ ($N$ particles and $M$ depth pixels) for each iteration. In our method, all particle fitness evaluation could be performed in parallel on the GPU (\Sec{impl}).
We have implemented our core algorithm in C++ and CUDA. Both the main optimization pipeline and the volumetric fusion run on a workstation with an Intel\textsuperscript{\textregistered} Core\textsuperscript{TM} i7-5930K CPU @ 3.50GHz $\times$ 12 with 32G RAM and an Nvidia GeForce RTX 2080 SUPER GPU with 8G memory. To enable flexible scanning, we implemented a front-end program running on a laptop for RGB-D capturing, compressing, and streaming to the workstation via WiFi.
Our pipeline runs with a framerate of $30$Hz for all shown test sequences. In fact, we control the time budget of random optimization process within $30$ms to maintain a $30$Hz framerate. This time slot allows the optimization to run for at least $20$ iterations which is well-sufficient for all the sequences tested as our method converges with $4$ iterations for ordinary sequences and $10$ for fast-motion ones. The time for volumetric depth fusion is $3$ms per frame.
\supl{The readers are welcomed to watch our accompanying video.}

\subsection{Limitation and Failure Cases}
\begin{figure}[t]
\centering
\begin{overpic}
[width=\linewidth]
{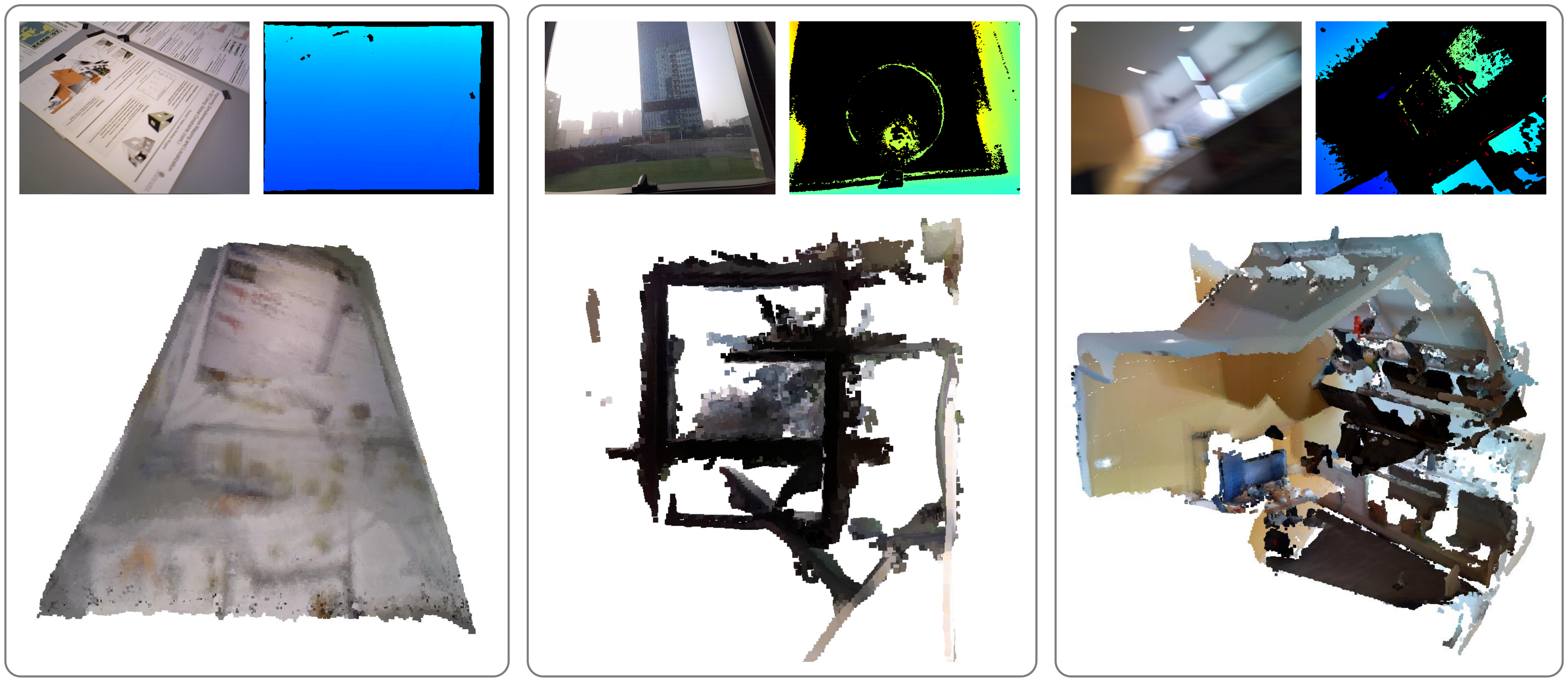}
\put(14,-3){\footnotesize (a)}
\put(48,-3){\footnotesize (b)}
\put(83,-3){\footnotesize (c)}
\end{overpic}
\caption{
Three typical failure cases of ROSEFusion: (a) almost no geometric feature (flat wall), (b) severe depth missing (transparent glass) and (c) too fast camera motion ($\sim 5$m/s).
}
\label{fig:failure-case}\vspace{-10pt}
\end{figure}

Our method has several limitations. \emph{Firstly}, since our method is purely geometrically-based, it would naturally fail when no prominent geometric feature can be found in a serial of consecutive frames. \Fig{failure-case}(a) shows such a cases where the camera is scanning towards a flat wall (\texttt{fr3\_nst} of \DSTUM). When the camera motion is not too fast to be contaminated by motion blur, one can always complement our method with photometric based pose optimization. \emph{Secondly}, when depth is severely missing due to reflectance or absorbance of light, our method would fail to track and fuse; see \Fig{failure-case}(b). \emph{Thirdly}, while it is hard to gauge the upper limit of camera moving speed, our method could still fail under extremely fast motion (e.g. $\sim 5$m/s in the example of \Fig{failure-case}(c)). \emph{Finally}, a more fundamental issue is that our current method does not include a loop closure detection and global pose optimization. Although we have tested scanning an indoor space of about $300$m$^2$ for $10$K frames without introducing observable drift, it is almost impossible to track without drifting when the sequence becomes extremely long especially some challenging frames were encountered along the way. The \texttt{Gym} sequence in \Fig{gallery} is a failure example of such kind. Nonetheless, our method can of course be enhanced by loop closure and/or global pose optimization.

\section{Discussion and conclusions}
\label{sec:future}
With our work, we wish to bring it to the community's attention that online dense reconstruction under fast camera motion is a practically useful problem and might demand a new paradigm of solution. The traditional approaches are mostly based on feature matching plus gradient descent. They do find difficulties in handling fast-motion camera tracking due to motion blur of color images and highly nonlinear/non-convex optimization landscape.

Our attempt achieves a satisfactory outcome thanks to two major design choices. \emph{First}, we pursue a depth-only approach and define an effective objective function based on depth-to-TSDF conformance, thus avoiding explicit feature detection and correspondence.
\emph{Second}, we propose to solve the per-frame optimization based on a novel particle-filter-based random optimization with swarm guidance, leading to a robust and efficient algorithm.

Our work serves merely as a starting point, which we hope would inspire a rich set future directions:
\begin{itemize}
  \item How to integrate loop closure detection and global pose optimization into the random optimization framework? It is interesting to study random optimization based pose graph optimization or bundle adjustment when large transformations are involved.
  \item How to unify traditional approach and random optimization into a single framework? When camera motion slows down, it is of course a good option to enhance camera tracking with photometric features. Here, how to bridge and switch smoothly between the two technique is worth of investigating.
  \item How to fuse multi-modal input to further improve the robustness of fast-motion camera tracking?
  \item How to realize autonomous reconstruction with mobile robots or drones with our technique? There are several technical issues to address, e.g., how to integrate our method with online motion planning~\cite{Xu15,xu2017autonomous,liu2018,dong2019multi}.
  \item It is interesting to extend our random optimization to distributed bundle adjustment~\cite{eriksson2016consensus,natesan2017distributed} for collaborative reconstruction by a network of robots or drones. Here, the small overlap between distributed nodes also calls for a robust pose optimization.
  \item Another promising and interesting direction is to explore the combination of random optimization and deep priors for online and/or semantic scene reconstruction~\cite{zhang2020fusion,huang2020di,nie2020rfd}.
\end{itemize} 

\begin{acks}
We thank the anonymous reviewers for their valuable comments and suggestions. We are grateful to Yuefeng Xi and Yao Chen for their effort in the preparation of the FastCaMo dataset. This work was supported in part by National Key Research and Development Program of China (2018AAA0102200), NSFC (61532003, 62002376) and NUDT Research Grants (ZK19-30).
\end{acks}

\bibliographystyle{ACM-Reference-Format}
\bibliography{fusion}

\end{document}